\documentclass{elsarticle}
\usepackage[utf8]{inputenc}
\usepackage[T1]{fontenc}
\usepackage[hidelinks]{hyperref}

\usepackage{tikz}
\usetikzlibrary{shapes,shapes.geometric,backgrounds,calc,fit,positioning}
\usepackage{amsmath, amssymb, amsfonts}
\newtheorem{definition}{Definition}
\usepackage{mathtools}
\usepackage{graphicx}
\usepackage[export]{adjustbox}
\usepackage{paralist}
\usepackage{textcomp}

\usepackage{booktabs}
\usepackage{arydshln} % other line variants
\usepackage{makecell}

\usepackage{float}
\usepackage{array}

\usepackage{xcolor}
\usepackage{fca}
\usepackage{nicefrac}
\usepackage{placeins}
\usepackage{todonotes}
\usepackage[toc,title,page]{appendix}

\usepackage{cleveref} 
\let\cref\Cref

\DeclareMathOperator{\supp}{supp}
\DeclareMathOperator{\conf}{conf}
\DeclareMathOperator{\Int}{Int}
\DeclareMathOperator{\Pclass}{P}

\newcommand{\limplies}{\rightarrow}

\let\citet\textcite

\newcommand{\SSHTM}{\texttt{SSH21}\xspace}
\DeclareMathOperator{\TM}{TM}
\DeclareMathOperator{\bow}{BoW}
\DeclareMathOperator{\tf}{tf}
\DeclareMathOperator{\idf}{idf}
\DeclareMathOperator{\tfidf}{tf-idf}
\newcommand{\Topics}{T}
\newcommand{\topic}{t}
\newcommand{\ntopics}{n}
\newcommand{\Documents}{D}
\newcommand{\documnt}{d}
\newcommand{\ndocuments}{l}
\newcommand{\Terms}{S}
\newcommand{\term}{s}
\newcommand{\nterms}{k}
\newcommand{\ITT}{\mathcal{S}}
\newcommand{\IDT}{\mathcal{D}}
\newcommand{\B}{\mathfrak{B}}
\newcommand{\abs}[1]{{\mid} #1 {\mid}}
\usepackage{colonequals}

\let\phi\varphi

\makeatletter
\tikzset{circle split part fill/.style  args={#1,#2}{%
 alias=tmp@name, 
  postaction={%
    insert path={
     \pgfextra{% 
     \pgfpointdiff{\pgfpointanchor{\pgf@node@name}{center}}%
                  {\pgfpointanchor{\pgf@node@name}{east}}%            
     \pgfmathsetmacro\insiderad{\pgf@x}
      \fill[#1] (\pgf@node@name.base) ([yshift=-\pgflinewidth]\pgf@node@name.north) arc
                          (90:270:\insiderad-\pgflinewidth)--cycle;
      \fill[#2] (\pgf@node@name.base) ([yshift=\pgflinewidth]\pgf@node@name.south)  arc
                           (270:450:\insiderad-\pgflinewidth)--cycle;            
         }}}}}  
\makeatother

\newcommand\diag[4]{
  \multicolumn{1}{|p{#2}||}{
    \hskip-\tabcolsep
    $\vcenter{\begin{tikzpicture}[baseline=0,anchor=south west,inner sep=#1]
        \path[use as bounding box] (0,0) rectangle (#2+2\tabcolsep,\baselineskip);
        \node[minimum width={#2+2\tabcolsep-\pgflinewidth},
        minimum  height=\baselineskip+\extrarowheight-\pgflinewidth] (box) {};
        \draw[line cap=round] (box.north west) -- (box.south east);
        \node[anchor=south west] at (box.south west) {#3};
        \node[anchor=north east] at (box.north east) {#4};
      \end{tikzpicture}}$\hskip-\tabcolsep}
}

\begin{document}

\begin{frontmatter}

\title{The Geometric Structure of Topic Models}

\author[1]{Johannes Hirth\corref{cor1}\fnref{fn1}}
\ead{hirth@cs.uni-kassel.de}
\cortext[cor1]{Corresponding author}

\affiliation[1]{organization={Knowledge \& Data Engineering Group, University of Kassel},
city={Kassel},
country={Germany}}

\author[2]{Tom Hanika\fnref{fn2}}
\ead{tom.hanika@uni-hildesheim.de}

\affiliation[2]{organization={Intelligent Information Systems, University of Hildesheim},
city={Hildesheim},
country={Germany}}

\begin{abstract}
  Topic models are a popular tool for clustering and analyzing textual
  data. They allow texts to be classified on the basis of their
  affiliation to the previously calculated topics. Despite their
  widespread use in research and application, an in-depth analysis of
  topic models is still an open research topic. State-of-the-art
  methods for interpreting topic models are based on simple
  visualizations, such as similarity matrices, top-term lists or
  embeddings, which are limited to a maximum of three dimensions.

  In this paper, we propose an incidence-geometric method for deriving
  an ordinal structure from flat topic models, such as non-negative
  matrix factorization.  These enable the analysis of the topic model
  in a higher (order) dimension and the possibility of extracting
  conceptual relationships between several topics at once.  Due to the
  use of conceptual scaling, our approach does not introduce any
  artificial topical relationships, such as artifacts of feature
  compression. Based on our findings, we present a new visualization
  paradigm for concept hierarchies based on ordinal motifs. These
  allow for a top-down view on topic spaces.
  
  We introduce and demonstrate the applicability of our approach based
  on a topic model derived from a corpus of scientific papers taken
  from 32 top machine learning venues.
\end{abstract}

\begin{keyword}
Publication Dynamics \sep Topic Models \sep Conceptual Views \sep Symbolic AI \sep Ordinal Data Science \sep Concept Hierarchies
\end{keyword}
\end{frontmatter}

\section{Introduction}
% Tom seed
% Topic Models natürlicherweise als Inzindenz Struktur zu verstehen und
% damit ist die Analyse deren Geometrie eine sinnvolle aufgabe.
\thispagestyle{empty}
Topic models are a popular tool for clustering and analyzing text
data. They discover groups or hierarchies of topics in text corpora
using various techniques. Using the computed topics one may classify
new text samples within the topic space. Furthermore, the underlying
vector space allows to enhance the understanding of the original text
corpus and the inter-topic relations of the documents. Thus, topic
models are an excellent tool for a structural analysis of large text
corpora. Moreover, using background information about corpus entities,
such as authors, venues, places, publishers, research groups, one may
extract topical knowledge about said entities.

The number of application domains for topic models is vast. Prominent
domains are recommendation systems~\cite{kawai2022topic,
  du2019cvtm,chen2020exploiting}, sentiment
analysis~\cite{VENUGOPALAN2022108668,zhou2022weakly,yin2022sentiment},
and text summarization~\cite{SRIVASTAVA2022108636,khanam2021joint,
  roul2021topic, frermann2019inducing}. A particular interesting
application is analyzing the temporal dynamics of corpora
entities~\cite{TopicSpaceTrajectories,MappingResearchTrajectories} and
creating maps from
them~\cite{daenekindt2020mapping,takizawa2023using}. These can also be
used for the holistic analysis of heterogeneous social network data,
such as co-authorship networks via text
corpora~\cite{schafermeier2022research,liao2020core}.

The vast majority of topic models encode relationships between topics
and terms (e.g., word or n-grams of these words). Hence, they are
comparatively easy to interpret in natural language.  This is in
particular true for the topic modeling procedure non-negative matrix
factorization (NMF) by D.\,D.~Lee and H.\,S.~Sung~\cite{nmf}, as it
enforces all components of a topic to be additive.

Common methods that aim at explaining and interpreting this relation
more formally rely on interpreting the elements of the underlying
vector space $\mathbb{R}^{t}$, where $t$ denotes the number of
topics. For example, in order to derive human-readable visualizations,
one does derive a two or three dimensional real-valued Euclidean space
from $\mathbb{R}^{t}$. However, the resulting plots are limited in
expressiveness by the employed maps from $\mathbb{R}^{t}$ to
$\mathbb{R}^{2}$. Even more serious, the information is potentially
distorted in the comparatively flat representations. This is due to
the non-linear nature of these maps. Therefore, these techniques
result in plots that are hard to interpret.

A fundamentially different line of research aims at explaining topic
models with methods rooted in ordinal data
analysis~\cite{cigarran2016step,6601864,castellanos2017formal,tods},
to which we want to contribute with the present work. For this purpose,
the term-topic and document-topic relations of a topic model are
implicitly understood as geometric incidence structures. With our work
we build on this view and make explicit use of the geometric
character. In detail, we show how these structures allow for rich
interpretations of the topic model and how to extract explainable
patterns from them. The latter are based on ordinal
motifs~\cite{ordinal-motif}, a novel approach for discovering
frequently occurring patterns in ordinal data. Moreover, we propose a
diagrammatic language for comprehensively visualizing the geometric
character of a topic model.

We demonstrate our approach based on a well researched topic model
that was derived from a large corpus of scientific works within the
realm of machine learning~\cite{TopicSpaceTrajectories}.  The results
show that our method is capable of capturing insights about authors
and research venues from the corpus data. In particular, our method is
capable of tracking temporal changes in individual topic distributions
of entities. Finally, we visually depict the interplay between terms
and their temporal dependencies for topics.

\section{Related Work}
Several methods to interpret topic models have been proposed.  Some of
them represent the document-topic relation through a vector space
models~\cite{TopicSpaceTrajectories,MappingResearchTrajectories}. This
allows the reader to visually interpret this relation through
proximity in two or three dimensional diagrams. Common methods for
this are multidimensional scaling \cite{MDS} or t-distributed
stochastic neighborhood embedding \cite{tsne}.

%% Relational explanations
Other explanation approaches follow a more relational approach.
% correlation 
A very simple method is to compute a relation based on topic-topic
correlation. This is, for example, computed using cosine
similarities between topics in a vector space model
\cite{TopicSpaceTrajectories}.
Some topic models, like the \emph{Correlation Topic Model} (CTM)
\cite{CTM}, allow for directly inferring this relation from the model
and presented in a graph structure. These approaches allow for
one-to-one comparisons of topics. Extensions to allow for hierarchical
interpretation of this relation has been proposed based on clustering
techniques \cite{CTM+FCA}.

% % topological
% Topic models based on \emph{topological data analysis} \todo{tda
% citation} allow for the extraction of binary topic relations based
% on topological properties \cite{Byrne2022TopicMW}.  

% incidence
With our work, we contribute to the explainability of topic models
based on relations that are defined on the data, i.e.,
documents-topics and term-topics relations.  These are often weighted
\cite{tfidf} and are \emph{scaled} to binary relations
\cite{scaling}. A basic investigation of the derived (binary)
relations are visualizations using bipartite graphs
\cite{crossno2011topicview}. Based on these graphs, simple (tree
shaped) hierarchies can be inferred by applying hierarchical
clustering methods \cite{Akhtar2018HierarchicalSO}. A benefit of
hierarchical interpretations is that one can infer topic-topic
relations of higher arity from them. On top of that, they allow for
assessing the overall \emph{global} structure of a data. However, tree
structures are very limited \cite{tree-view} in their
expressiveness. For instance, there is only a single path connection
two nodes.

% FCA
A hierarchical structure that is not limited by this property can be
computed using \emph{Formal Concept Analysis}~\cite{fca} (FCA). With
FCA, we compute from the bipartite graph the set of all maximal
bi-cliques. These exhibit a natural order relation which results in a
lattice structure, called the \emph{concept lattice}. An application
of concept lattices on documents-topics and term-topics relations has
been shown to be useful for organizing discussion forums in
educational software \cite{TM+FCA}. We further elaborate on this
method after an introduction to FCA in \cref{sec:fca}.
With our work, we expand on FCA based methods by transferring new
state-of-the-art~\cite{ordinal-motif,ordinal-motif-covering}
explanation approaches for concept lattice into the realm of topic
modeling. These allow for a novel global interpretation of the topic
model structure.

% Hierarchical topic models
A different, yet related line of research is the computation of
hierarchical topic models
\cite{FCA-TM2,zhao2018inter,hlda,PAM-HTM}. Multiple visualization
techniques have been proposed for their explanation
\cite{tmhierarchy}.

\section{Topic Models and their Interpretations}
\emph{Topic Models} are unsupervised machine learning procedure that
identifies \emph{abstract topics} in a collection of documents.  The
literature proposes plenty techniques for that task.  Starting with a
set of \emph{textual} documents $\Documents$ (\emph{corpus}) most
approaches compute a vector representation for all the \emph{terms}
$\Terms$ (e.g. words) occurring in the corpus.  The two most commonly
used instances of vector representations are \emph{bag-of-words} (BoW)
and \emph{tf-idf}. The BoW model represents each document by a tuple
in $\mathbb{N}^{\Terms}$ that reflects for each term $\term\in \Terms$
the number occurrences in $\documnt$. Since all consecutive
computations are conducted in a real-valued $\abs{\Terms}$-dimensional
vector space, the tuple is identified with a vector in said space.
Formally, for the set of all terms $\Terms$ in the corpus $\Documents$
a document $\documnt\in \Documents$ is mapped to $\mathbb{R}^{\abs{\Terms}}$,
where $\bow(\documnt)\coloneqq (v_1,\dots,v_\nterms)$ and $v_i$ is
equal to the number of occurrences of $\term_i$ in $\documnt$.

The tf-idf representation consecutively adds a measure of importance
per term. This measure reflects the rarity of terms in the corpus in
combination with the absolute frequency in the document.
More formally, the \emph{inverted document frequency} (idf) of a term
$\term\in \Terms$ in $\Documents$ equals the logarithm of the inverse
of the document occurrences of $\term$, i.e.,
$\idf(\term;\Documents)\coloneqq \log(
\nicefrac{\abs{\Documents}}{\abs{\{\documnt\in \Documents\mid \term\in
    \documnt\}}})$.
The \emph{term frequency} (tf) of a term $\term$ in a document $\documnt$
is equal to the normalized BoW representation of $\term$ in
$\documnt$, i.e.,
$\tf(\term;\documnt)\coloneqq
\nicefrac{\bow(\documnt)_{\term}}{\abs{\bow(\documnt)}}$ where
$\bow(\documnt)_{\term}$ equals the value of $\bow(\documnt)$ for term
$\term$ and $\abs{\bow(\documnt)}$ equals the sum of all values. 
Combined, the tf-idf is defined as
$\tfidf(\documnt)\coloneqq
\big(\tfidf(\term_1;\documnt,\Documents),\dots,\tfidf(\term_\nterms;\documnt,\Documents)\big)$
where is defined as
$\tfidf(\term_1;\documnt,\Documents)\coloneqq \tf(\term;\documnt)\cdot
\idf(\term;\Documents)$. 

Based on this representation of a corpus one may compute a topic model
$\TM$. Most frequently, the result is a relation between terms and
found topics. By using this relation, each document can be mapped into
$\mathbb{R}^{\ntopics}$, where $\ntopics$ is the number of
topics. Vectors of this space can encode any share and combination of
topics. To make this \emph{topic space} more comprehensible,
$\mathbb{R}^{\ntopics}$ is often substituted by $[0,1]^{\ntopics}$.
This is usually done by normalizing vectors through various methods.
Since topics in a topic model are comprised of combinations of terms,
they allow for human comprehension to some extent.

\subsection{Topic Model Visualization}
% \begin{figure}
%   \centering
%   \includegraphics[width=0.35\linewidth]{pics/nips.png}
%   \includegraphics[width=0.35\linewidth]{pics/RecSys.png}
%   \includegraphics[width=0.35\linewidth]{pics/neuralnetworks.png}
%   \includegraphics[width=0.35\linewidth]{pics/ecml.png}
%   \includegraphics[width=0.9\linewidth]{pics/topic_ids.png}
%   \caption{From Topic Space Trajectories Figure 4.6}
% \end{figure}

\begin{figure}
  \centering
  \includegraphics[trim=155 0 155 0,clip, width=0.95\linewidth]{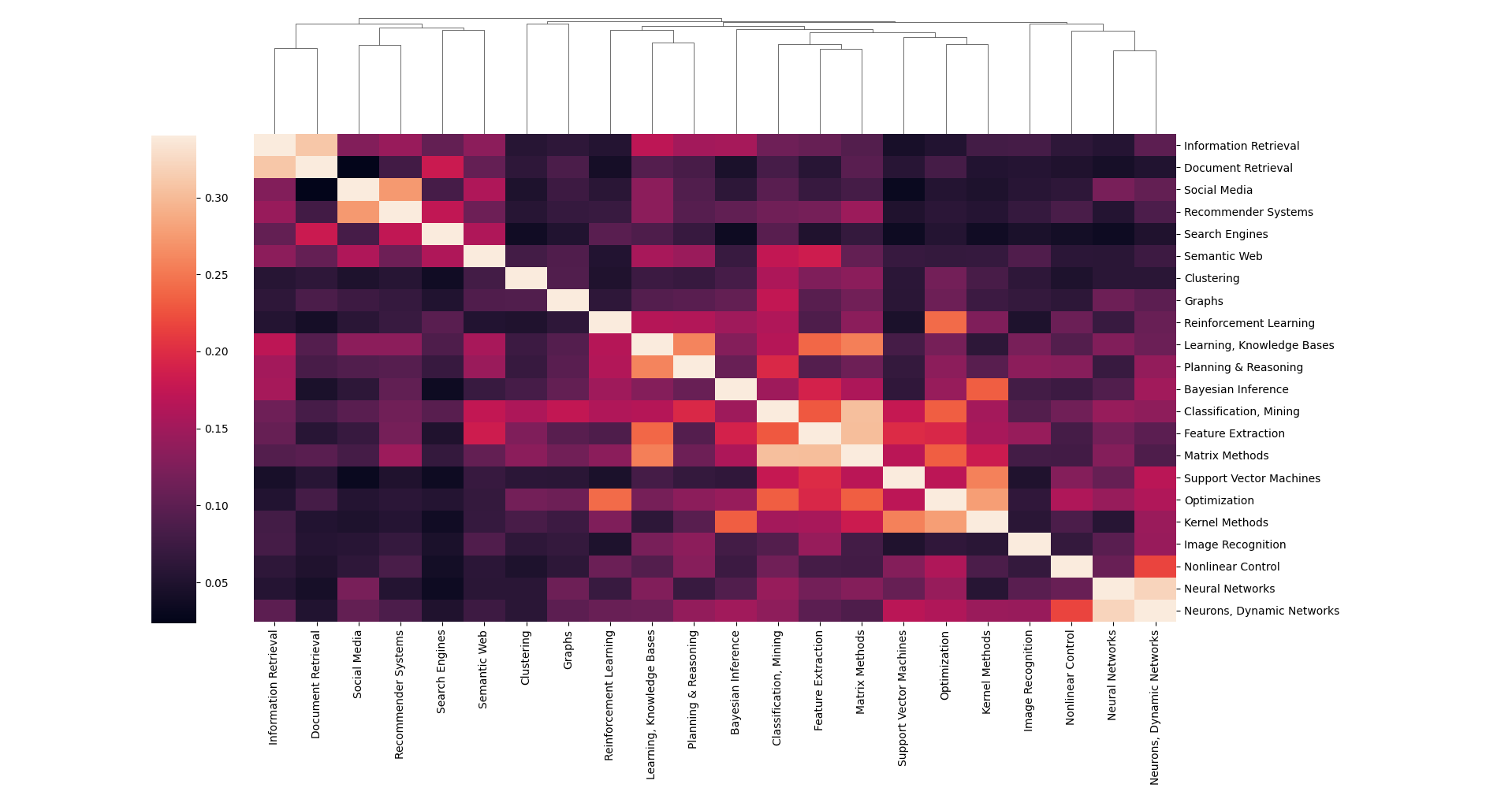}
  \caption{The similarity heatmap of the topics from the \SSHTM topic model 
    \cite[Figure 4.1]{TopicSpaceTrajectories}.}
  \label{fig:topic-model1}
\end{figure}

\begin{figure}
  \centering
  \includegraphics[ width=0.4\linewidth]{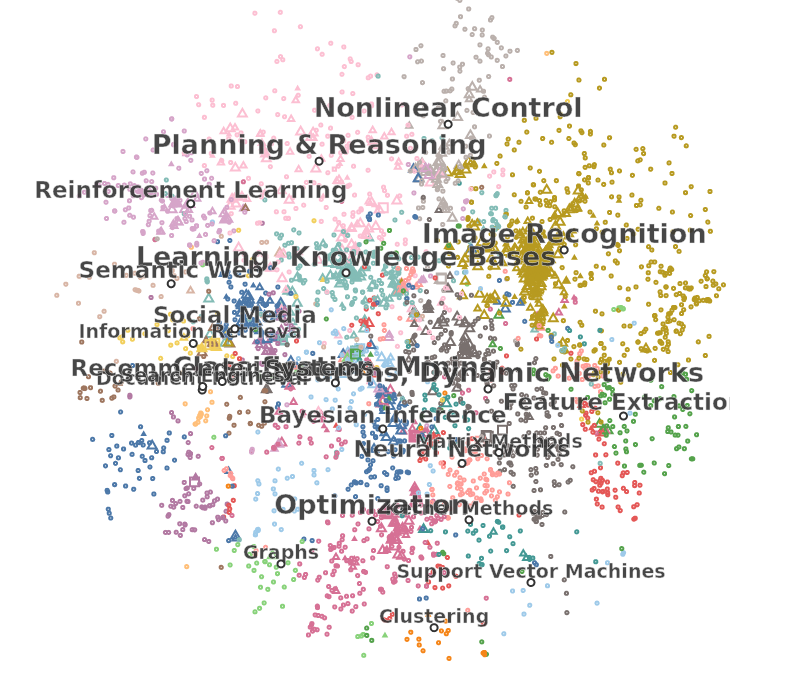}
  \includegraphics[ width=0.59\linewidth]{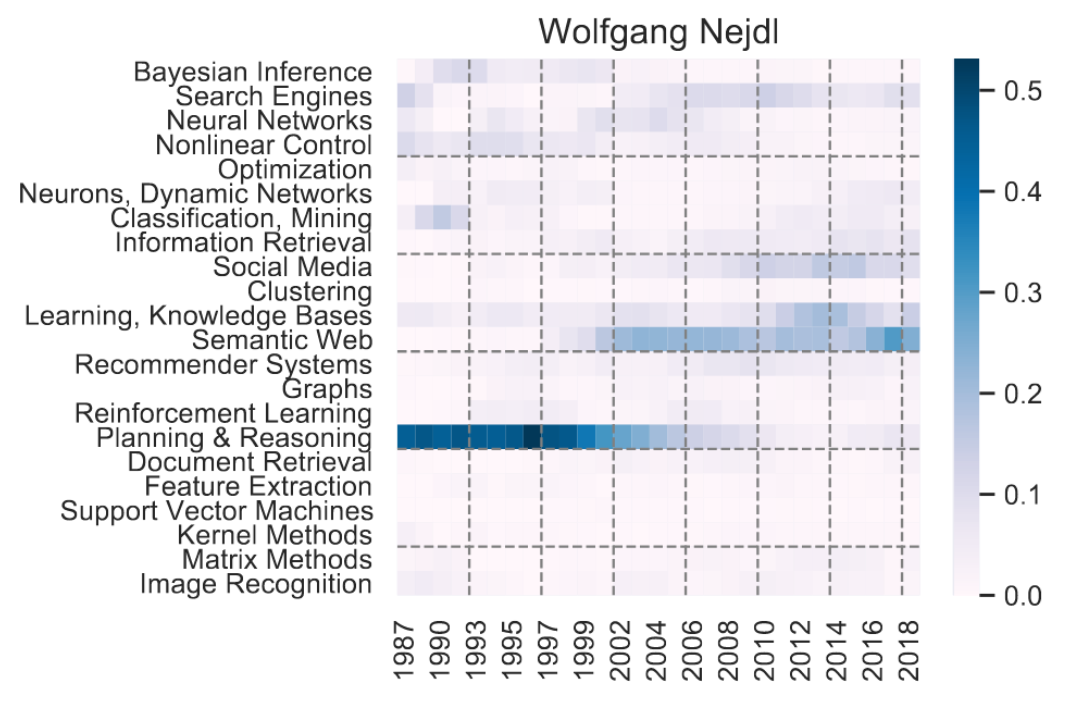}
  \caption{Visualizations of the \SSHTM topic model from the
    literature. The similarity heatmap (top) is from Figure 4.1 in
    Schäfermeier et al.~\cite{TopicSpaceTrajectories}, 
    The vector space representation 
    \cite[\url{https://sci-rec.org/maps}]{MappingResearchTrajectories} and the heatmap from 
    Figure 2 in \cite{MappingResearchTrajectories} of the \SSHTM topic model.}
  \label{fig:topic-model2}
\end{figure}

With our work, we contribute towards a rich interpretation of
documents in the topic space. We compare our method to three commonly
used explanation techniques for topic models. Such methods are usually
based on visualizations. We present how they can be used based on the
well understood \SSHTM topic model from
Schäfermeier et al.\cite{TopicSpaceTrajectories}. This topic model was computed on
machine learning research papers \cite{SemanticScholarGraph} and has
twenty-two topics. The topic model itself is discussed in more detail
in \cref{chapter:topic:sec:view}.
 
% cosine sim
The first visualization of \SSHTM, depicted in
\cref{fig:topic-model1}, is a similarity heatmap. This plot displays
in each cell the output of a similarity measure between two
topics. This similarity is computed using the term-topic relations
discussed in \cref{chapter:topic:sec:relation}.  From this plot one
can infer the one-to-one topic relations between topics and
potentially identify small clusters of topics.  Yet, it is difficult
to infer n-to-n topic relations.  Moreover, this plot does not reveal
the reasons for similarities and dissimilarities of topics. Therefore,
this method eludes from explainability.  We consider it as a necessity
to give explanations of such (dis-) similarities in the language of the
terms.

% embedding
Another commonly used visualization of topic models are embeddings
into the $\mathbb{R}^{2}$ or (rarely) the $\mathbb{R}^{3}$ vector
space, e.g., umap \cite{umap} or t-SNE \cite{tsne}. These
visualizations aim to reflect similarity through (local) proximity of
topics. In \cref{fig:topic-model2} (left) we depict an embedding of
the \SSHTM topic model (from Schäfermeier et
al.~\cite{MappingResearchTrajectories}) computed with t-SNE. The
resulting visualizations are considered good with respect to the
application domains \cite{tsne-capable}.  However, because of the
distortion of the original distances by non-linear mappings it is very
difficult to relate the distance within the t-SNE plot to meaningful
distances between topics.  This makes it difficult to assess
topic-topic relations from the resulting diagrams.  This problem is
amplified if clusters are computed based on the distorted
distances. For obvious reasons, approaches employing linear mappings
into $\mathbb{R}^{2}$ do often fail to separate classes
\cite{tsne-capable} or lack performance \cite{embeddings-comparison}.
Although this approach to analyze topic models is more geometric in
nature, it does not correctly reflect the incidences between topics
and terms. 

% time plot
The third visualization technique maps the topic representation of an
entity in regards to temporal information. These entities can be
collections of papers representing authors or venues. Every cell in
this heatmap (cf. \cref{fig:topic-model2}, right) reflects the topic
share of an entity for a particular year. For example, in the depicted
picture, we find that \emph{Wolfgang Nejdl's} research focus in 2017
was on \emph{Semantic Web}. Moreover, such plots allow to identify
changes in topics over time.  Although this visualization technique is
promising for analyzing single entities, it lacks geometric depth. For
example, it is unclear if in 2004 all documents were about
\emph{Semantic Web} and \emph{Planning \& Reasoning}, or if these sets
were disjoint. A different view on this is the question, to what
extent both topics are related. We deem an answer to this question a
necessity for a comprehensive topic analysis method.

In the following, we introduce our novel method for analyzing and
explaining topic models based on an (order-)relational
approach. Alongside a practical investigation, we revisit the just
discussed techniques and compare them to our results.

\section{Conceptual Views on Topic Models}\label{sec:bv_analysis}
In contrast to the methods in the last section, from now on, we want
to discuss hierarchical approaches. Eventually, we will focus on
concept based methods rooted in the theory of Formal Concept Analysis
(FCA) \cite{fca,Wille-lattice}. For this we require a proper definition of the
incidences that naturally result from topic models. Based on this we
will review previous research on topic modeling with FCA. We will
extend some of these approaches in the next main section and
consecutively develop a novel theory for representing and visualizing
topic models on a global scale.

\subsection{Incidence Relations in Topic Models}\label{chapter:topic:sec:relation}
\begin{figure}
  \setlength{\extrarowheight}{0.1cm}
  \newcolumntype{x}[1]{>{\centering\arraybackslash}p{#1}}
  \begin{minipage}{.49\linewidth}
    \centering
    \begin{tabular}{|x{0.5cm}||x{0.5cm}|x{0.5cm}|x{0.5cm}|x{0.5cm}|}
      \hline
      \diag{.1em}{.5cm}{$\Documents$}{$\Topics$}&$\topic_1$&$\topic_2$&$\dots$&$\topic_\ntopics$ \\ \hline \hline
      $\documnt_1$&$w_{1,1}$&$w_{1,2}$&&$w_{1,\ntopics}$\\ \hline
      $\documnt_2$&$w_{2,1}$&$w_{2,2}$&&$w_{2,\ntopics}$\\ \hline
      $\dots$&&&&\\ \hline
      $\documnt_\ndocuments$&$w_{\ndocuments,1}$&$w_{\ndocuments,2}$&&$w_{\ndocuments,\ntopics}$\\ \hline
    \end{tabular}

  \end{minipage}
  \begin{minipage}{.49\linewidth}
    \centering
    \begin{tabular}{|x{0.5cm}||x{0.5cm}|x{0.5cm}|x{0.5cm}|x{0.5cm}|}
      \hline
      \diag{.1em}{.5cm}{$\Terms$}{$\Topics$}&$\topic_1$&$\topic_2$&$\dots$&$\topic_\ntopics$ \\ \hline \hline

      $\term_1$&$\hat{w}_{1,1}$&$\hat{w}_{1,2}$&&$\hat{w}_{1,\ntopics}$\\ \hline
      $\term_2$&$\hat{w}_{2,1}$&$\hat{w}_{2,2}$&&$\hat{w}_{2,\ntopics}$\\ \hline
      $\dots$&&&&\\ \hline
      $\term_\nterms$&$\hat{w}_{\nterms,1}$&$\hat{w}_{\nterms,2}$&&$\hat{w}_{\nterms,\ntopics}$\\ \hline
    \end{tabular}
  \end{minipage}
  \caption{The weighted term-topic (left) and document-topic relations (right).}
  \label{fig:topic-term-matrix}
\end{figure}

The goal of this section is to present a principled approach to
extract relational structures from topic models. Every topic
model $\TM$ exhibits at least two fundamental relations, i.e., the
topics for any document $\documnt\in \Documents$ and the terms for a given
topic $\topic\in \Topics$.  Often the relations of $\TM$ are weighted,
e.g., a document might have topic $\topic_1$ with a weight of $w\in [0,1]$.
We depict these relations in \cref{fig:topic-term-matrix} via a
document topic matrix (left) and a topic term matrix (right). 
The first matrix represents the documents in a lower dimensional topic
space.  The second matrix provides an interpretation of the topic
space and is used for state of the art topic model evaluation measures
\cite{topic-model-coherence-validation,tm-eval} like
\texttt{npmi}~\cite{coherencemodel}.

The document topic matrix can be extracted by embedding each document
$\documnt\in \Documents$ by the topic model $\TM$ into the topic space, i.e.,
computing $\TM(\documnt_i)= (w_{i,1},\dots,w_{i,\ntopics})$. For the computation of
the term topic matrix there are multiple options depending on the used
topic model method. The first is to embedd a document that is only
composed of term $\topic_j$ into the topic space. This results in a row in
the term topic matrix. The second option is applicable to methods that
have an additional decoder map from the topic space to the document
space, e.g., auto-encoder or NMF. In this case we can map the vector
$(0,0,\dots,1,\dots,0)$ that has a zero for all topics but $t_j$ by
the decoder. The result is a column in term topic matrix. 

Based on both matrices we can infer incidence relations, i.e., binary
relations $\IDT \subseteq \Documents \times \Topics$ and
$\ITT \subseteq \Terms \times \Topics$, in the following way. One
natural approach is to apply threshold values to the weights.  We
apply this to the document topic matrix by selecting a value
$\delta \in [0,1]$. This results in the
\emph{document-topic-incidence} $\IDT_{\delta}$ with
$(d,t)\in \IDT_{\delta}$ iff the weight for $(\documnt,\topic)$ in the document
topic matrix is greater than or equal to $\delta$. For the
\emph{term-topic-incidence} $\ITT$ pursue a different path. We extract
for each topic the top-n terms, i.e., the top $n$ entries in the
respective column in the term topic matrix. 
For a term $\term\in \Terms$ and a topic $\topic\in \Topics$ is
$(\term,\topic)\in \ITT_{n}$ iff the weight of $(\term,\topic)$ is among the top-n
greatest weights in column $\topic$ in the term topic matrix.

\subsection{Formal Concept Analysis}\label{sec:fca}
An extensive mathematical toolset to analyze the resulting incidence
relations is Formal Concept Analysis~\cite{fca}. It allows to
cluster elements from the incidence in a human-comprehensible
way. Moreover, these clusters form an hierarchical structure in a
natural way. Most importantly, FCA is equipped with a rich toolset to
interpret the resulting hierarchies, which is rooted in human
conceptual thinking \cite{ConceptualStructures,whatAreConcepts}. 

The basic data structure in FCA is the \emph{formal context}, i.e., a
triple $\context \coloneqq (G,M,I)$ where $G$ is a set called
\emph{objects}, $M$ is a set called \emph{attributes} and
$I\subseteq G\times M$, called incidence. We interpret $(g,m)\in I$ as
\emph{object $g$ has attribute $m$}. From the incidence follow two
natural maps, the \emph{object derivation}
$(\cdot)':\mathcal{P}(G)\to \mathcal{P}(M)$ with
$A'\coloneqq \{m\in M\mid \forall g\in A: (g,m)\in I\}$, and the
\emph{attribute derivation}
$(\cdot)':\mathcal{P}(M)\to \mathcal{P}(G)$ where
$B'\coloneqq \{g\in G\mid \forall m\in B: (g,m)\in I\}$. We use for
both maps the same symbol by abuse of notation.

A pair $(A,B)\in \mathcal{P}(G)\times \mathcal{M}$ with $A'=B$ and
$A=B'$ is called a \emph{formal concept} of $\context$. The set $A$
can be understood as cluster of objects of $G$ and $B$ its description
in terms of attributes. The set of all formal concepts of $\context$
is denoted $\B(\context)$. Concepts are ordered by the
\emph{subconcept} relation $\leq$ where $(A,B)\leq (C,D)$ iff
$A\subseteq C$. Hence, the pair
$\BV(\context)\coloneqq (\B(\context),\leq)$ constitutes an ordered
set. More precisely $\BV(\context)$ is lattice ordered, i.e., for any
set of concepts $\mathcal{C}\subseteq \BV(\context)$ we can compute
their greatest lower bound (meet) as well as their least upper bound
(join). Therefor, $\BV(\context)$ is called \emph{concept lattice}.
Altogether, $\BV(\context)$ allows us to derive an hierarchical
structure for documents, terms and topics from topic models. 

The state-of-the-art on analyzing topic models with
FCA is to derive a formal context from the document-topic $\IDT$ and
the term-topic relation $\ITT$ \cite{TM+FCA}. This is done by applying
thresholds to both relations as discussed in
\cref{chapter:topic:sec:relation}. However, for the term-topic
relation $\ITT$, we use the top-$n$ term relation, since topic model
evaluation measures based on this relation, like npmi
\texttt{npmi}~\cite{coherencemodel}, correlate with human
interpretation of topics
\cite{topic-model-coherence-validation,tm-eval}. So far, the resulting
concept lattices have mainly been used as a means to navigate between
forum entries. The interpretation of the topic model with respect to
the concept lattice is to the best of our knowledge not studied.

\section{Conceptual Views}\label{chapter:topic:sec:view}
We now extend the just introduced procedures by means of a novel
approach called \emph{ordinal
  motifs}~\cite{ordinal-motif,ordinal-motif-covering}. These motifs
are user defined ordinal substructures that allow for a rich
interpretation of the underlying data. It is a similar approach to the
analysis of hierarchical structures as network motifs are for
analyzing (social) network graphs.

We want to introduce and demonstrate our novel approach based on an
already published and extensively discussed topic
model~\cite{TopicSpaceTrajectories} which we call in the following
\SSHTM.
It was build based on a document corpus on 35,200 scientific
publications from the realm of machine learning research. It was
consecutively evaluated on a corpus of about 350,000 documents from
the same domain. All documents were retrieved via the \emph{Semantic
  Scholar Open Research Corpus}~\cite{SemanticScholarGraph}. The
employed topic modeling technique is \emph{non-negative matrix
  factorization}~\cite{nmf}, a widely used procedure that has several
advantages with respect to explainability . We may remark at this
point that our notion is agnostic with respect to the topic modeling
technique.  The topic model consists of twenty-two~\cite[Table
4.1]{TopicSpaceTrajectories} topics which were manually assigned based
on the top ten terms per topic. The training corpus and therefor the
resulting topic model has 14828 terms.

\subsection{Computing Incidence Relations}\label{chapter:topic:sec:relation}

\begin{figure}
  \centering
  \begin{minipage}{.45\linewidth}
    \includegraphics[width=1\linewidth]{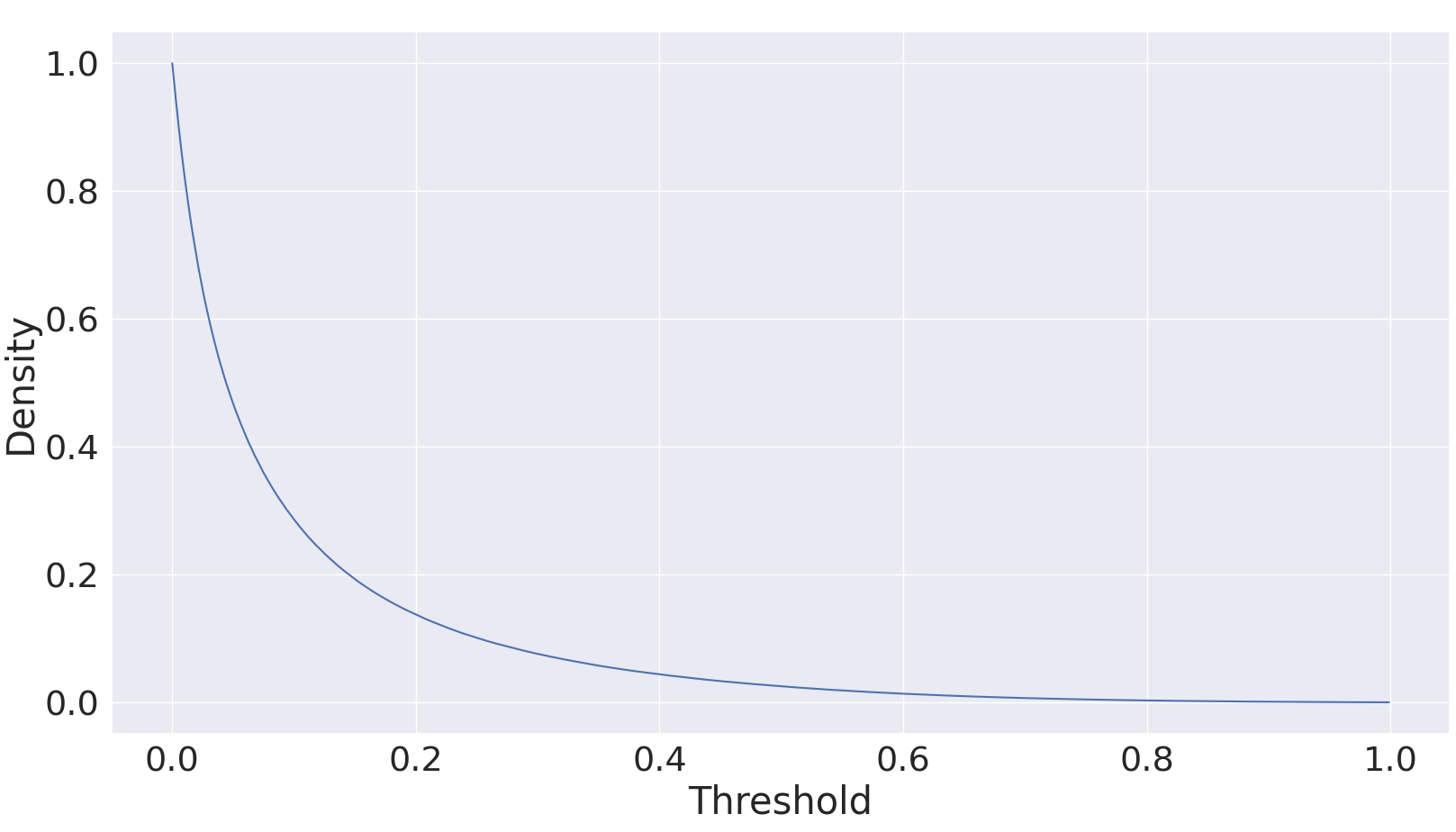}    
  \end{minipage}\hfill
  \begin{minipage}{.51\linewidth}
    \tabcolsep=2pt\scriptsize
    \begin{tabular}{|l||r|r|r|r|r|r|}
    \hline
    & 0.1 & 0.2 & 0.25 & 0.3 & 0.4 & 0.5\\
    \hline\hline
    Sebastian Thrun    & 398 &  89 & 51   & 31  & 18  & 13 \\
    Bernhard Schölkopf & 640 & 144 & 80   & 51  & 20  & 17 \\
    Dieter Fox         & 342 & 66  & 42   & 28  & 15  & 10 \\
    Wolfgang Nejdl     & 387 & 94  & 58   & 39  & 19  & 14 \\
      \hline
ECML           &  568 & 173 &  108 &  74 &  37 &  20\\
RecSys         &  389 &  83 &   60 &  47 &  21 &  11\\
NeurIPS        & 2366 & 405 &  212 & 142 &  66 &  22\\
Neural Networks& 1368 & 261 &  155 & 107 &  43 &  21\\
\hline
    \end{tabular}

  \end{minipage}
  \caption{The density of the document-topic relation for given
    thresholds (left) and the concept lattice sizes for given entities
    and thresholds (right).}
  \label{fig:scaling}
\end{figure}

For computing the document-topic-incidence $\IDT$ we have to set a
threshold $\delta$. Since the respective document topic matrix is (or
can be) row normalized we have to choose a value from $[0,1]$.  The
goal for any choice of $\delta$ is to derive a sparse \cite{sparseTM}
document topic incidence. Thereby documents are mainly represented by
their most important topics. Moreover, this leads to a comprehensibly
sized concept lattice, which fosters the overall understanding of the
results. However, at the same time increasing the values for $\delta$
to much may lead to loosing substantial parts of the concept lattice
structure.  We decided to determine $\delta$ based on the resulting
density of $\IDT$. That is
$\abs{\IDT}/\abs{\Documents \times \Topics}$, which is depicted on the
left in \cref{fig:scaling}.

We also want to propose a threshold estimation method tailored for
investigating particular entities of a document corpus, such as
authors or publication venues. This requires background knowledge
about the topic corpus, e.g., which documents belong to a particular
author or which documents were published at a certain venue.  To
address the implicit goal for achieving a comprehensible number of
formal concepts in these cases, we also computed their number for
selected values of $\delta$ and four different authors as well as four
different venues, see \cref{fig:scaling} right. Computing the number
of concepts for a particular scientist $a$ or venue $v$ means that we
considered only documents that were co-authored by $a$ or published at
$v$. The result is an induced sub-context of the $\IDT_{\delta}$
formal context. For the set of venues we decided to look into
\emph{ECML, RecSys, NeurIPS} and \emph{Neural Networks} as they were
extensively discussed for the \SSHTM\cite{TopicSpaceTrajectories}. As
for the set of authors we chose \emph{Thrun, Schölkopf, Fox} and
\emph{Nejdl} since their individual publication trajectories were
extensively discussed in a follow-up
paper~\cite{MappingResearchTrajectories}.

Based on the results that we achieved and reported in
\cref{fig:scaling}, we decided for the threshold $\delta=0.25$. We
acknowledge that the number of formal concepts is still high in some
cases. We depict the resulting number of concepts as well as the sizes
of the induced subcontexts in the first four columns of
\cref{fig:scaling}.

\begin{table}
  \centering
  \caption{The table displays the number of objects, attributes, the
    density and the number of concepts of the context derived from
    $\IDT$ for four authors and four venues. The sixth column depicts
    the number of concepts in the $2,8$-core and the last column the
    number of concepts after applying TITANIC with a minimum support
    value of three percent.}
  \tabcolsep=3pt\scriptsize
  \begin{tabular}{|l||r|r|r|r|r|r|}
    \hline
                        & \parbox{1.6cm}{\centering documents (objects)} & \parbox{1.7cm}{\centering topics \\(attributes)} & density & concepts& \parbox{1.4cm}{\centering core\\ concepts} & \parbox{1.4cm}{\centering view\\ concepts}\\
    % view concepts are from core+titanic                       
    \hline\hline
    Sebastian Thrun    & 266 & 22 & 0.055 & 51 & 11&11\\
    Bernhard Schölkopf & 522 & 22 & 0.059 & 80 & 50&22\\
    Dieter Fox         & 244 & 22 & 0.057 & 42 & 16&15\\
    Wolfgang Nejdl     & 387 & 22 & 0.059 & 58 & 28&21\\
      \hline
        ECML           & 639 & 22 & 0.061 & 108  & 52&25\\
        RecSys         & 856 & 22 & 0.058 & 60   & 29&23 \\
        NeurIPS        & 6233 & 22 & 0.060 & 212 & 202&25\\
        Neural Networks& 3521 & 22 & 0.061 & 155 & 144&21\\
\hline
    \end{tabular}
  \label{fig:view-stats}
\end{table}

Formal Concept Analysis has a rich tool-set of data reduction
methods. A particular feature of these tools is that they allow for
controlling the (conceptual)
error~\cite{smeasure,smeasure-error,smeasure-explore}.  The threshold
for the $\ITT$ will be discussed in
\cref{chapter:topic:sec:term-topic-view}.

\subsection{Conceptual Data Reduction}\label{chapter:topic:sec:reduction}
In order to reduce the size of the just computed incidence relations
we rely on two established methods from FCA, TITANIC~\cite{titanic}
and $pq$-cores~\cite{pqcores}. The overall goal is to compute
hierarchical representations of comprehensible size.
We consider diagrams of size up to thirty or in some cases up to fifty
concepts to have diagrams that are presentable in a human
comprehensible way.

\paragraph{$pq$-cores}
The technique $pq$-cores computes the densest part of $\IDT$. That is,
the largest subset of documents $H\subseteq \Documents$ and topics
$S\subseteq \Topics$ such that for each document $d\in H$ has at least
$p$ topics and every topic $t\in S$ has at least $q$ documents in
$\IDT$. This method can easily be restricted to a proper subset of the
documents, such as all documents belonging to an author $a$ or a venue
$v$. We then call the result the \emph{core topics} of an author or a
venue respectively. 

The proper choice of parameters $p,q$ is supported by an importance
measure.  A parameter pair is considered to be interesting with
respect to the data if every increase in $p$ or $q$ causes a large
reduction in the number of formal concepts. In our study this lead to
the pair $p=2$ and $q=8$.

\paragraph{TITANIC}
The \texttt{TITANIC} algorithm computes the hierarchy of formal
concepts in a top down fashion, with respect to a pre-defined
importance parameter.  For this parameter one can choose the value of
a monotonous function on the set of concept intents. That is
monotonous with respect to set inclusion. A commonly used function for
this task is the support function, i.e.,
$\supp_{\context}:\Int(\context)\to [0,1]$, where
$\supp(B)\coloneqq \nicefrac{\abs{B'}}{\abs{G}}$. In other words, the
support of an intent $B$ reflects the relative number of objects that
have all attributes from $B$. 
Based on this, the \texttt{TITANIC} algorithm computes the hierarchy
of all formal concepts that satisfy a minimum threshold value
$c\in [0,1]$. The result is called iceberg concept lattice, i.e., a
\emph{join-semilattice}. 
The main advantage of the \texttt{TITANIC} algorithm is that it
computes concept hierarchies of readable size.

In our case study on the \SSHTM topic model we found the value
$c=0.03$ to be sufficient for in \cref{chapter:topic:sec:relation}
computed sub-contexts.

The reduction in terms of formal concepts by both methods are reported
in the last two columns of \cref{fig:view-stats}.

\subsection{The Resulting Conceptual View and Interpretation}\label{chapter:topic:sec:view-interpretation}

\begin{figure}[!ht]
\hspace{-1cm}\scalebox{0.9}{\input{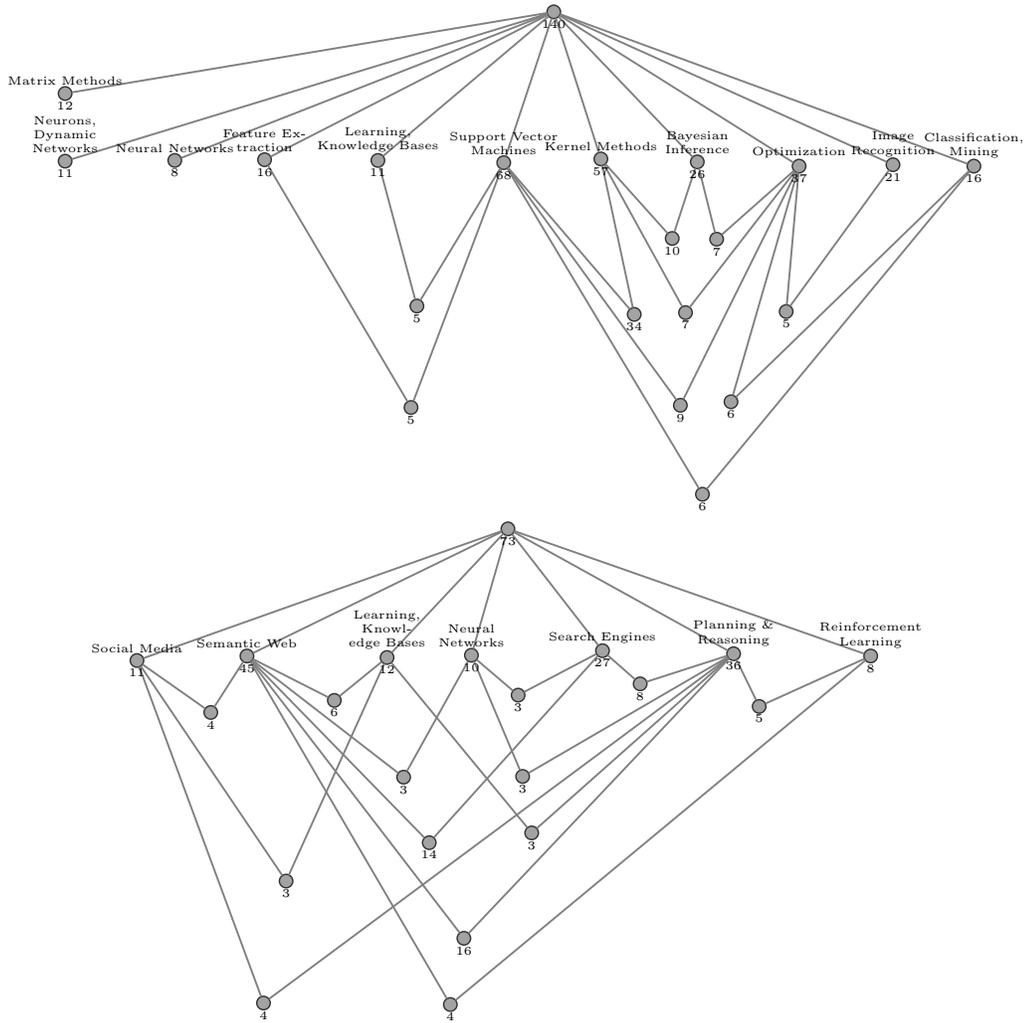}}
\scalebox{0.9}{\colorlet{mivertexcolor}{black!80}
\colorlet{jivertexcolor}{black!80}
\colorlet{vertexcolor}{black!80}
\colorlet{bordercolor}{black!80}
\colorlet{linecolor}{gray}
% parameter corresponds to the used valuation function and can be addressed by #1
\tikzset{vertexbase/.style 2 args={semithick, shape=circle, inner sep=2pt, outer sep=0pt, draw=bordercolor},%
  vertex/.style 2 args={vertexbase={#1}{}, fill=vertexcolor!45},%
  mivertex/.style 2 args={vertexbase={#1}{}, fill=mivertexcolor!45},%
  jivertex/.style 2 args={vertexbase={#1}{}, fill=jivertexcolor!45},%
  divertex/.style 2 args={vertexbase={#1}{}, top color=mivertexcolor!45, bottom color=jivertexcolor!45},%
  conn/.style={-, thick, color=linecolor}%
}
\tikzstyle{d} = [text width=1.7cm,align=center]
\begin{tikzpicture}[yscale=0.1,xscale=0.2,font=\tiny]
  \begin{scope} %for scaling and the like
    \begin{scope} %draw vertices
      \foreach \nodename/\nodetype/\param/\xpos/\ypos in {%
        0/vertex//47.99701997459976/-0.9026477947164295,%["2004" "2006" "2004" "2009"]
        1/vertex//34.19871313295423/-0.680094458560859,%["2002" "2003" "1998" "2015"]
        2/vertex//48.99850998729983/8.889698996128786,%["2002" "2004" "2005" "2004" "2004" "2007" "2003" "2001" "2007" "2001" "2002" "1998" "2001" "2003" "2008" "2001"]
        3/vertex//35.86786315412103/17.346725770040564,%["2003" "2013" "2004"]
        4/vertex//46.43914662151075/23.021835842007675,%["2016" "2006" "2006" "2007" "2008" "2010" "2009" "2005" "2004" "2009" "2010" "2016" "2012" "2009"]
        5/vertex//54.00596005080023/24.46843252701889,%["2002" "2003" "2002"]
        6/vertex//44.54744326418838/32.7029059647751,%["2002" "2002" "2002"]
        7/vertex//53.33830004233351/32.81418263285289,%["2004" "2003" "1994"]
        8/vertex//30.3040297502317/42.272699419464736,%["2009" "2010" "1998" "2012"]
        9/vertex//70.80873693054599/43.16291276408703,%["2007" "2004" "1996" "2009" "1993"]
        10/vertex//39.4287165326102/44.05312610870932,%["2006" "2009" "2016" "2003" "2000" "2009"]
        11/vertex//53.00447003810016/44.832062785253825,%["2006" "2004" "2005"]
        12/vertex//62.017880152400856/46.501212806420625,%["2004" "2004" "1997" "2008" "2008" "1987" "2009" "2004"]
        13/mivertex//24.851473014420172/49.95078951683201,%["2009" "2002" "2012" "2003" "2010" "1998" "2013" "2012" "2015" "2004" "2005"]
        14/mivertex//43.32339991533272/50.39589618914315,%["2006" "2002" "2009" "2003" "2009" "2016" "2013" "2003" "2000" "2004" "2009" "2002"]
        15/mivertex//32.97466978409859/50.61844952529873,%["2002" "2016" "2004" "2006" "2005" "2009" "2004" "2004" "2006" "2007" "2006" "2007" "2008" "2003" "2001" "2010" "2007" "2006" "2009" "2005" "2004" "2009" "2016" "2001" "2002" "2010" "2002" "2002" "1998" "2012" "2003" "2009" "2000" "2001" "2009" "2004" "2003" "2002" "2009" "2010" "2008" "2016" "2001" "2012" "2009"]
        16/mivertex//79.0432103683022/50.61844952529873,%["2007" "2004" "2006" "1996" "2009" "2004" "2009" "1993"]
        17/mivertex//49.55489332768877/50.729726193376514,%["2004" "2003" "2006" "2004" "2002" "2002" "2002" "1994" "2005" "2005"]
        18/mivertex//68.91703357322362/50.952279529532085,%["2004" "2002" "2004" "2005" "2004" "2004" "2007" "1997" "2004" "2004" "2002" "2007" "2003" "2002" "2003" "2001" "2003" "2007" "2008" "1996" "2008" "2009" "2001" "2002" "1998" "2001" "2015" "1993" "1987" "2003" "2008" "2001" "1994" "2009" "2002" "2004"]
        19/mivertex//59.235963450456204/51.39738620184323,%["2004" "2016" "2004" "1997" "2006" "2012" "2006" "2007" "2008" "2009" "2010" "2006" "2008" "2004" "2009" "2005" "2008" "1987" "2004" "2009" "2010" "2016" "2009" "2005" "2012" "2009" "2004"]
        20/vertex//52.25081961885415/69.41111883772174%["2004" "2002" "2016" "2004" "2006" "2005" "2004" "2004" "2007" "2009" "1997" "2004" "2004" "2002" "2006" "2007" "2012" "2006" "2003" "2002" "2007" "2008" "2003" "2009" "2001" "2010" "2003" "2006" "2007" "2008" "2006" "1996" "2004" "2009" "2005" "2008" "2009" "2004" "2009" "2016" "2001" "2002" "2010" "2002" "2002" "1998" "2013" "2012" "2003" "2009" "2000" "2001" "2015" "2004" "1993" "1987" "2009" "2004" "2003" "2002" "2009" "2010" "2008" "2016" "2001" "1994" "2009" "2005" "2012" "2002" "2009" "2005" "2004"]
      } \node[\nodetype={\param}{}] (\nodename) at (\xpos, \ypos) {};
    \end{scope}
    \begin{scope} %draw connections
      \path (12) edge[conn] (19);
      \path (1) edge[conn] (13);
      \path (5) edge[conn] (14);
      \path (17) edge[conn] (20);
      \path (14) edge[conn] (20);
      \path (9) edge[conn] (18);
      \path (19) edge[conn] (20);
      \path (13) edge[conn] (20);
      \path (6) edge[conn] (15);
      \path (10) edge[conn] (15);
      \path (8) edge[conn] (15);
      \path (15) edge[conn] (20);
      \path (11) edge[conn] (19);
      \path (6) edge[conn] (17);
      \path (3) edge[conn] (13);
      \path (11) edge[conn] (17);
      \path (8) edge[conn] (13);
      \path (7) edge[conn] (18);
      \path (18) edge[conn] (20);
      \path (2) edge[conn] (15);
      \path (1) edge[conn] (18);
      \path (4) edge[conn] (19);
      \path (9) edge[conn] (16);
      \path (16) edge[conn] (20);
      \path (3) edge[conn] (14);
      \path (10) edge[conn] (14);
      \path (5) edge[conn] (18);
      \path (0) edge[conn] (16);
      \path (12) edge[conn] (18);
      \path (0) edge[conn] (15);
      \path (2) edge[conn] (18);
      \path (7) edge[conn] (17);
      \path (4) edge[conn] (15);
    \end{scope}
    \begin{scope} %add labels
      \foreach \nodename/\labelpos/\labelopts/\labelcontent in {%
        13/above/d/{Social Media},
        14/above/d/{Learning, Knowledge Bases},
        15/above/d/{Semantic Web},
        16/above/d/{Reinforcement Learning},
        17/above/d/{Neural Networks},
        18/above/d/{Planning \& Reasoning},
        19/above/d/{Search Engines},
        0/below//{4},
        1/below//{4},
        2/below//{16},
        3/below//{3},
        4/below//{14},
        5/below//{3},
        6/below//{3},
        7/below//{3},
        8/below//{4},
        9/below//{5},
        10/below//{6},
        11/below//{3},
        12/below//{8},
        13/below//{11},
        14/below//{12},
        15/below//{45},
        16/below//{8},
        17/below//{10},
        18/below//{36},
        19/below//{27},
        20/below//{73},
        0/right//{},%["2004" "2006" "2004" "2009"]
        1/right//{},%["2002" "2003" "1998" "2015"]
        2/right//{},%["2002" "2004" "2005" "2004" "2004" "2007" "2003" "2001" "2007" "2001" "2002" "1998" "2001" "2003" "2008" "2001"]
        3/right//{},%["2003" "2013" "2004"]
        4/right//{},%["2016" "2006" "2006" "2007" "2008" "2010" "2009" "2005" "2004" "2009" "2010" "2016" "2012" "2009"]
        5/right//{},%["2002" "2003" "2002"]
        6/right//{},%["2002" "2002" "2002"]
        7/right//{},%["2004" "2003" "1994"]
        8/right//{},%["2009" "2010" "1998" "2012"]
        9/right//{},%["2007" "2004" "1996" "2009" "1993"]
        10/right//{},%["2006" "2009" "2016" "2003" "2000" "2009"]
        11/right//{},%["2006" "2004" "2005"]
        12/right//{},%["2004" "2004" "1997" "2008" "2008" "1987" "2009" "2004"]
        13/right//{},%["2009" "2002" "2012" "2003" "2010" "1998" "2013" "2012" "2015" "2004" "2005"]
        14/right//{},%["2006" "2002" "2009" "2003" "2009" "2016" "2013" "2003" "2000" "2004" "2009" "2002"]
        15/right//{},%["2002" "2016" "2004" "2006" "2005" "2009" "2004" "2004" "2006" "2007" "2006" "2007" "2008" "2003" "2001" "2010" "2007" "2006" "2009" "2005" "2004" "2009" "2016" "2001" "2002" "2010" "2002" "2002" "1998" "2012" "2003" "2009" "2000" "2001" "2009" "2004" "2003" "2002" "2009" "2010" "2008" "2016" "2001" "2012" "2009"]
        16/right//{},%["2007" "2004" "2006" "1996" "2009" "2004" "2009" "1993"]
        17/right//{},%["2004" "2003" "2006" "2004" "2002" "2002" "2002" "1994" "2005" "2005"]
        18/right//{},%["2004" "2002" "2004" "2005" "2004" "2004" "2007" "1997" "2004" "2004" "2002" "2007" "2003" "2002" "2003" "2001" "2003" "2007" "2008" "1996" "2008" "2009" "2001" "2002" "1998" "2001" "2015" "1993" "1987" "2003" "2008" "2001" "1994" "2009" "2002" "2004"]
        19/right//{},%["2004" "2016" "2004" "1997" "2006" "2012" "2006" "2007" "2008" "2009" "2010" "2006" "2008" "2004" "2009" "2005" "2008" "1987" "2004" "2009" "2010" "2016" "2009" "2005" "2012" "2009" "2004"]
        20/right//{}%["2004" "2002" "2016" "2004" "2006" "2005" "2004" "2004" "2007" "2009" "1997" "2004" "2004" "2002" "2006" "2007" "2012" "2006" "2003" "2002" "2007" "2008" "2003" "2009" "2001" "2010" "2003" "2006" "2007" "2008" "2006" "1996" "2004" "2009" "2005" "2008" "2009" "2004" "2009" "2016" "2001" "2002" "2010" "2002" "2002" "1998" "2013" "2012" "2003" "2009" "2000" "2001" "2015" "2004" "1993" "1987" "2009" "2004" "2003" "2002" "2009" "2010" "2008" "2016" "2001" "1994" "2009" "2005" "2012" "2002" "2009" "2005" "2004"]
      } \coordinate[label={[\labelopts]\labelpos:{\labelcontent}}](c) at (\nodename);
    \end{scope}
  \end{scope}
\end{tikzpicture}}
  \caption{The concept lattice for the entities B.~Schölkopf (top) and W.~Nejdl (bottom).}
  \label{fig:schoelkopf-nejdl-view}
\end{figure}

In \cref{fig:schoelkopf-nejdl-view} we depict the iceberg concept
lattice for the researcher Bernhard Schölkopf\footnote{\url{https://dblp.org/pid/97/119.html}}
$\BV_{\text{BS}}$ (top) and Wolfgang Nejdl\footnote{\url{https://dblp.org/pid/n/WolfgangNejdl.html}}
$\BV_{\text{WN}}$ (bot). We employ order diagrams with a particular
\emph{short-hand} notation as follows: Each concept is represented by
a node in the diagram. For example the node annotated with
\emph{Support Vector Machines} is the concept $(A,B)$ that collects
all documents that share the topic
$B=\{\emph{Support Vector Machines}\}$. In classical FCA one annotates
all these documents below the node. Due to their large quantity, we
annotated only their number, i.e., $\abs{A}=68$.

The connection between two nodes indicates that the corresponding
concepts are in subconcept relation. For example, the node in the
bottom right annotated by $6$ is a subconcept of the concept with the
intent \emph{Support Vector Machines}. It is also a subconcept of the
concept with the intent \emph{Classification, Mining}. The extent size
6 indicates that there are six documents in $\IDT$ of Bernhard
Schölkopf that have the topics \emph{Support Vector Machines} and
\emph{Classification, Mining}.  In general, the larger concept in the
subconcept relation is placed higher in the diagram and the topic
names (attributes) are annotated atop the largest concept that they
are contained in.  The top most concept bears no annotation. This
indicates that there is no topic that is shared by all documents.

As a first observation, we can read from $\B_{\text{WN}}$ that only
seven topics out of twenty-two were identified as core topics of
Wolfgang Nejdl by the combined method.  The most frequent topic is
\emph{Semantic Web} which occurs in forty-five documents.
Out of these, sixteen also have the \emph{Planning \& Reasoning} topic
and fourteen are also associated to \emph{Search Engines}.  The second
most frequent topic is \emph{Planning \& Reasoning}, which occurs in
thirty-six documents. Overall, the $\B_{\text{WN}}$
iceberg concept lattice has twenty-one concepts.

This novel approach for a comprehensive analysis of a topic model with
respect to an entity allows for several new insights.

By combining the core approach with TITANIC, we identified those
topics for an entity that are not only frequent but also strongly
interconnected~\cite{pqcores}. With this structural approach, we
overcome the limitations of commonly used methods that are based on
filtering topics by frequency. For example, selecting the strongest
signals in \cref{fig:topic-model1,fig:topic-model2} would result in a
total ordered ranking without structural insights. In particular, one
cannot infer how the topics are connected in terms of shared
documents.  Another advantage of our approach is that infrequent and
isolated topics are omitted.

We present the same analysis for $\BV_{\text{BS}}$. For Bernhard
Schölkopf we can identify twenty-two concepts comprised of eleven core
topics. The most supported and structurally important topics are
\emph{Support Vector Machines} with sixty-eight documents
and \emph{Kernel Methods} with fifty-seven documents.
The topics coincide in thirty-four documents.
The large overlap of these topics is not surprising due to the close
connection between SVMs and kernel methods.

Apart from the individual analysis of an entity's topic structure, our
novel approach does also enable for an in-depth cross entity
comparisons. Comparing $\BV_{\text{BS}}$ and $\BV_{\text{WN}}$ we
observe that both authors have two topics in common, namely
\emph{Learning Knowledge Bases} and \emph{Neural Networks}.

Despite that, these topics are completely differently interconnected
within their respective research. While Wolfgang Nejdl studies
\emph{Learning Knowledge Bases} in the context of \emph{Semantic Web,
  Social Media} and \emph{Planning \& Reasoning}, Bernhard Schölkopf
studies \emph{Learning Knowledge Bases} in the context of
\emph{Support Vector Machines}. An analog differentiation can be found
for \emph{Neural Networks}.

\textbf{Ordinal Motifs in Lattices:}\quad %
The geometric aspects of our analysis method allow the
application of \emph{ordinal motifs}.  This method on how to interpret
concept hierarchies was proposed by Hirth et al.~\cite{ordinal-motif}
and is capable of explaining concept lattices
\cite{ordinal-motif-covering} based on common scales, i.e., ordinal
patterns.  In detail, this method extracts sub-structures within the
concept hierarchy and provides visual \emph{geometric}
interpretations.  We discuss the three types of ordinal motifs that
occurred in our data. It is quite possible that for other data sets
other ordinal motifs might occur. Nonetheless, our method can be
applied analogously.

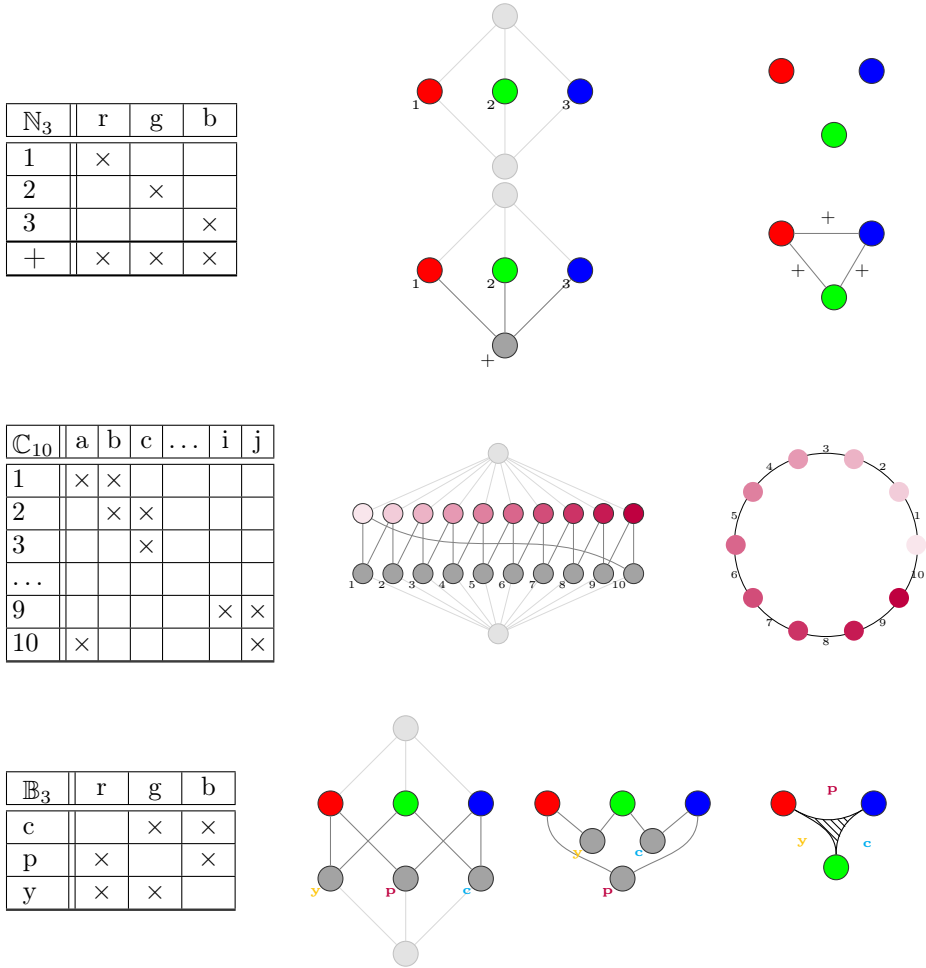
\begin{figure}[t!]
  \begin{minipage}[t][5cm][c]{1.0\linewidth}
  \begin{minipage}{0.28\linewidth}
    \begin{tabular}{|l||c|c|c|}
      \hline
      \multicolumn{1}{|c||}{\raisebox{-1.5pt}{$\mathbb{N}_3$}}&r&g&b\\ \hline \hline
      1&$\times$&&\\ \hline
      2&&$\times$&\\ \hline
      3&&&$\times$\\ 
      \Xhline{2\arrayrulewidth}
      \textbf{+}&$\times$&$\times$&$\times$\\ 
      \Xhline{2\arrayrulewidth}
    \end{tabular}
  \end{minipage}
  % \begin{minipage}[b][1.75cm][c]{0.42\linewidth}
  %   \begin{minipage}[c][1cm][t]{.49\linewidth}
  %     \begin{cxt} 
  %       \cxtName{$\mathbb{N}_3$}
  %       \att{r}
  %       \att{g}
  %       \att{b}
  %       \obj{x..}{1}
  %       \obj{.x.}{2}
  %       \obj{..x}{3}
  %     \end{cxt}
  %   \end{minipage}
  %   \begin{minipage}[c][1cm][t]{.49\linewidth}
  %     \begin{cxt} 
  %       \cxtName{$\mathbb{N}^{+}_3$}
  %       \att{r}
  %       \att{g}
  %       \att{b}
  %       \obj{x..}{1}
  %       \obj{.x.}{2}
  %       \obj{..x}{3}
  %       \obj{xxx}{\textbf{+}}
  %     \end{cxt}
  %   \end{minipage}
  % \end{minipage}
  \begin{minipage}{0.02\linewidth}\,
  \end{minipage}
  \begin{minipage}{0.44\linewidth}\centering
    \begin{tikzpicture}[concept/.style={fill=black!36,shape=circle,draw=black!80},
      hidden/.style={opacity=0.3},
      edge/.style={color=gray},]
      % bot                      
      \node[concept,hidden] (a) at (0,0){};
      % row 1
      \node[concept,fill=red] (b) at (-1,1){};
      \node[concept,fill=green] (c) at (0,1){};
      \node[concept,fill=blue] (d) at (1,1){};
      % top
      \node[concept,hidden] (h) at (0,2){};
      % bot to row 1
      \path[edge,hidden] (a) edge (b);
      \path[edge,hidden] (a) edge (c);
      \path[edge,hidden] (a) edge (d);
      % row 2 to top
      \path[edge,hidden] (b) edge (h);
      \path[edge,hidden] (c) edge (h);
      \path[edge,hidden] (d) edge (h);

      \coordinate[label={below left:{\tiny 1}}]() at (b);
      \coordinate[label={below left:{\tiny 2}}]() at (c);
      \coordinate[label={below left:{\tiny 3}}]() at (d);
    \end{tikzpicture}

    \begin{tikzpicture}[concept/.style={fill=black!36,shape=circle,draw=black!80},
      hidden/.style={opacity=0.3},
      edge/.style={color=gray}]
      
      \node[concept,hidden] (top) at (0,3) {};

      \node[concept,fill=red] (a) at (-1,2) {};
      \node[concept,fill=green] (b) at (0,2) {};
      \node[concept,fill=blue] (c) at (1,2) {};

      \node[concept] (bot) at (0,1) {};

      \path[edge,hidden] (top) edge (a);
      \path[edge,hidden] (top) edge (b);
      \path[edge,hidden] (top) edge (c);

      \path[edge] (a) edge (bot);
      \path[edge] (b) edge (bot);
      \path[edge] (c) edge (bot);

      \coordinate[label={below left:{\tiny 1}}]() at (a);
      \coordinate[label={below left:{\tiny 2}}]() at (b);
      \coordinate[label={below left:{\tiny 3}}]() at (c);

      \coordinate[label={below left:{\tiny \textbf{+}}}]() at (bot);
    \end{tikzpicture}
  \end{minipage}
  \begin{minipage}{0.2\linewidth}\centering
    \begin{minipage}[b][1.74cm][c]{.99\linewidth}
      \begin{tikzpicture}[concept/.style={fill=black!36,shape=circle,draw=black!80},
        hidden/.style={opacity=0.5},
        edge/.style={color=gray},]
        % bottom limit                      
        \node[concept,opacity=0] (a) at (0,0){};

        % cube nodes
        \node[concept,fill=red] (b) at (0.8,2){};
        \node[concept,fill=green] (c) at (1.5,1.15){};
        \node[concept,fill=blue] (d) at (2,2){};

        % top limit
        \node[concept,opacity=0] (a) at (0,3){};
      \end{tikzpicture}      
    \end{minipage}
    \begin{minipage}[t][1.74cm][c]{.99\linewidth}
      \begin{tikzpicture}[concept/.style={fill=black!36,shape=circle,draw=black!80},
        hidden/.style={opacity=0.3},
        edge/.style={color=gray}]

        % bottom limit                      
        \node[concept,opacity=0] (a) at (0,0){};

        % cube nodes
        \node[concept,fill=red] (b) at (0.8,2){};
        \node[concept,fill=green] (c) at (1.5,1.15){};
        \node[concept,fill=blue] (d) at (2,2){};

        % top limit
        \node[concept,opacity=0] (a) at (0,3){};

        % round cube edges
        \path[edge] (b) edge (c);
        \path[edge] (b) edge (d);
        \path[edge] (c) edge (d);

        \node[opacity=0.0] (a) at (0,3) {};

        \coordinate[label={above left:{\tiny \textbf{+}}}]() at (1.65,2);
        \coordinate[label={above left:{\tiny \textbf{+}}}]() at (1.25,1.3);
        \coordinate[label={above left:{\tiny \textbf{+}}}]() at (2.1,1.3);

        % \coordinate[label={above left:{\tiny b}}]() at (b);
        % \coordinate[label={above left:{\tiny c}}]() at (c);
        
        % Check if needed?
        % \coordinate[label={below left:{\tiny 1}}]() at (b);
        % \coordinate[label={below left:{\tiny 2}}]() at (c);
        % \coordinate[label={below right:{\tiny 3}}]() at (d);
        
        % \coordinate[label={below left:{\tiny \textbf{+}}}]() at (bot);
      \end{tikzpicture}
    \end{minipage}
  \end{minipage}
\end{minipage}
%%% Local Variables:
%%% mode: latex
%%% TeX-master: "../paper"
%%% End:

  \begin{minipage}[t][3.5cm][c]{1.0\linewidth}
  \setlength\tabcolsep{2pt}
  \begin{minipage}{0.28\linewidth}
  \begin{cxt} 
    \cxtName{$\mathbb{C}_{10}$}
    \att{a}
    \att{b}
    \att{c}
    \att{$\dots$}
    \att{i}  
    \att{j}  
    \obj{xx....}{1}
    \obj{.xx...}{2}
    \obj{..x...}{3}
    \obj{......}{$\dots$}
    \obj{....xx}{9}
    \obj{x....x}{10}
  \end{cxt}    
  \end{minipage}
  \begin{minipage}{0.48\linewidth}\centering
    \scalebox{0.8}{
      \begin{tikzpicture}[concept/.style={fill=black!36,shape=circle,draw=black!80},
                      hidden/.style={opacity=0.3},
                      edge/.style={color=gray}]
    % bot                      
    \node[concept,hidden] (a) at (0,0){};
    % row 1
    \node[concept] (ba) at (-2.25,1){};
    \node[concept] (bb) at (-1.75,1){};
    \node[concept] (bc) at (-1.25,1){};
    \node[concept] (bd) at (-0.75,1){};
    \node[concept] (be) at (-0.25,1){};
    \node[concept] (bf) at (0.25,1){};
    \node[concept] (bg) at (0.75,1){};
    \node[concept] (bh) at (1.25,1){};
    \node[concept] (bi) at (1.75,1){};
    \node[concept] (bj) at (2.25,1){};
    % row 1 labels
    \coordinate[label={below left:{\tiny 1}}](c) at (ba);
    \coordinate[label={below left:{\tiny 2}}](c) at (bb);
    \coordinate[label={below left:{\tiny 3}}](c) at (bc);
    \coordinate[label={below left:{\tiny 4}}](c) at (bd);
    \coordinate[label={below left:{\tiny 5}}](c) at (be);
    \coordinate[label={below left:{\tiny 6}}](c) at (bf);
    \coordinate[label={below left:{\tiny 7}}](c) at (bg);
    \coordinate[label={below left:{\tiny 8}}](c) at (bh);
    \coordinate[label={below left:{\tiny 9}}](c) at (bi);
    \coordinate[label={below left:{\tiny 10}}](c) at (bj);

    % row 2
    \node[concept,fill=purple!10] (ca) at (-2.25,2){};
    \node[concept,fill=purple!20] (cb) at (-1.75,2){};
    \node[concept,fill=purple!30] (cc) at (-1.25,2){};
    \node[concept,fill=purple!40] (cd) at (-0.75,2){};
    \node[concept,fill=purple!50] (ce) at (-0.25,2){};
    \node[concept,fill=purple!60] (cf) at (0.25,2){};
    \node[concept,fill=purple!70] (cg) at (0.75,2){};
    \node[concept,fill=purple!80] (ch) at (1.25,2){};
    \node[concept,fill=purple!90] (ci) at (1.75,2){};
    \node[concept,fill=purple!100] (cj) at (2.25,2){};

    % top
    \node[concept,hidden] (h) at (0,3){};
    % bot to row 1
    \path[edge,hidden] (a) edge (ba);
    \path[edge,hidden] (a) edge (bb);
    \path[edge,hidden] (a) edge (bc);
    \path[edge,hidden] (a) edge (bd);
    \path[edge,hidden] (a) edge (be);
    \path[edge,hidden] (a) edge (bf);
    \path[edge,hidden] (a) edge (bg);
    \path[edge,hidden] (a) edge (bh);
    \path[edge,hidden] (a) edge (bi);
    \path[edge,hidden] (a) edge (bj);

    % row 1 to row 2
    \path[edge] (ba) edge (ca);
    \path[edge] (bb) edge (cb);
    \path[edge] (bc) edge (cc);
    \path[edge] (bd) edge (cd);
    \path[edge] (be) edge (ce);
    \path[edge] (bf) edge (cf);
    \path[edge] (bg) edge (cg);
    \path[edge] (bh) edge (ch);
    \path[edge] (bi) edge (ci);
    \path[edge] (bj) edge (cj);
    
    \path[edge] (ba) edge (cb);
    \path[edge] (bb) edge (cc);
    \path[edge] (bc) edge (cd);
    \path[edge] (bd) edge (ce);
    \path[edge] (be) edge (cf);
    \path[edge] (bf) edge (cg);
    \path[edge] (bg) edge (ch);
    \path[edge] (bh) edge (ci);
    \path[edge] (bi) edge (cj);
    \path[edge,in=-30,out=150] (bj) edge (ca);

    % row 2 to top
    \path[edge,hidden] (h) edge (ca);
    \path[edge,hidden] (h) edge (cb);
    \path[edge,hidden] (h) edge (cc);
    \path[edge,hidden] (h) edge (cd);
    \path[edge,hidden] (h) edge (ce);
    \path[edge,hidden] (h) edge (cf);
    \path[edge,hidden] (h) edge (cg);
    \path[edge,hidden] (h) edge (ch);
    \path[edge,hidden] (h) edge (ci);
    \path[edge,hidden] (h) edge (cj);
  \end{tikzpicture}}
  \end{minipage}
  \begin{minipage}{0.22\linewidth}\hfill
    \scalebox{0.8}{
  \begin{tikzpicture}[edge/.style={color=gray}]
\node[shape=circle,scale=9.2,draw=black] at (0,0){};

    % for a in range(0,10):
    %    angle = a*2*math.pi/10
    %    coords = (math.cos(angle)*1.5,math.sin(angle)*1.5)
    %    print(f"\\node[fill=purple!{10*a + 10}] ({a}) at {coords}{'{}'};")
\node[shape=circle,fill=purple!10] (0) at (1.5, 0.0){};
\node[shape=circle,fill=purple!20] (1) at (1.2135254915624212, 0.8816778784387097){};
\node[shape=circle,fill=purple!30] (2) at (0.4635254915624212, 1.4265847744427302){};
\node[shape=circle,fill=purple!40] (3) at (-0.463525491562421, 1.4265847744427305){};
\node[shape=circle,fill=purple!50] (4) at (-1.213525491562421, 0.8816778784387098){};
\node[shape=circle,fill=purple!60] (5) at (-1.5, 1.8369701987210297e-16){};
\node[shape=circle,fill=purple!70] (6) at (-1.2135254915624214, -0.8816778784387096){};
\node[shape=circle,fill=purple!80] (7) at (-0.46352549156242134, -1.4265847744427302){};
\node[shape=circle,fill=purple!90] (8) at (0.46352549156242084, -1.4265847744427305){};
\node[shape=circle,fill=purple!100] (9) at (1.213525491562421, -0.88167787843871){};

\node[] () at (1.5216904260722457, 0.49442719099991583){\tiny 1};
\node[] () at (0.940456403667957, 1.294427190999916){\tiny 2};
\node[] () at (9.797174393178826e-17, 1.6){\tiny 3};
\node[] () at (-0.9404564036679569, 1.294427190999916){\tiny 4};
\node[] () at (-1.5216904260722457, 0.49442719099991606){\tiny 5};
\node[] () at (-1.521690426072246, -0.49442719099991567){\tiny 6};
\node[] () at (-0.9404564036679572, -1.2944271909999159){\tiny 7};
\node[] () at (-2.9391523179536476e-16, -1.6){\tiny 8};
\node[] () at (0.9404564036679567, -1.294427190999916){\tiny 9};
\node[] () at (1.5216904260722457, -0.4944271909999162){\tiny 10};
  \end{tikzpicture}}
  \end{minipage}
\end{minipage}
%%% Local Variables:
%%% mode: latex
%%% TeX-master: "../paper"
%%% End:

  \begin{minipage}[t][3.5cm][c]{1.0\linewidth}
  \begin{minipage}{0.28\linewidth}
    \begin{cxt} 
      \cxtName{$\mathbb{B}_3$}
      \att{r}
      \att{g}
      \att{b}
      \obj{.xx}{c}
      \obj{x.x}{p}
      \obj{xx.}{y}
  \end{cxt}
  \end{minipage}
  \begin{minipage}{0.28\linewidth}\centering
  \begin{tikzpicture}[concept/.style={fill=black!36,shape=circle,draw=black!80},
                      hidden/.style={opacity=0.3},
                      edge/.style={color=gray},]
    % bot                      
    \node[concept,hidden] (a) at (0,0){};
    % row 1
    \node[concept] (b) at (-1,1){}; %,fill=yellow!75!orange
    \node[concept] (c) at (0,1){}; %,fill=purple
    \node[concept] (d) at (1,1){};%,fill=cyan
    % row 2
    \node[concept,fill=red] (e) at (-1,2){};
    \node[concept,fill=green] (f) at (0,2){};
    \node[concept,fill=blue] (g) at (1,2){};
    % top
    \node[concept,hidden] (h) at (0,3){};
    % bot to row 1
    \path[edge,hidden] (a) edge (b);
    \path[edge,hidden] (a) edge (c);
    \path[edge,hidden] (a) edge (d);
    % row 1 to row 2
    \path[edge] (b) edge (e);
    \path[edge] (b) edge (f);
    \path[edge] (c) edge (e);
    \path[edge] (c) edge (g);
    \path[edge] (d) edge (f);
    \path[edge] (d) edge (g);
    % row 2 to top
    \path[edge,hidden] (e) edge (h);
    \path[edge,hidden] (f) edge (h);
    \path[edge,hidden] (g) edge (h);

    \coordinate[label={below left:{\color{yellow!55!orange} \bfseries\tiny y}}]() at (b);
    \coordinate[label={below left:{\color{purple} \bfseries \tiny p}}]() at (c);
    \coordinate[label={below left:{\color{cyan} \bfseries \tiny c}}]() at (d);
  \end{tikzpicture}
  \end{minipage}
  \begin{minipage}{0.18\linewidth}\centering
  \begin{tikzpicture}[concept/.style={fill=black!36,shape=circle,draw=black!80},
                      hidden/.style={opacity=0.3},
                      edge/.style={color=gray},]
    % bot limit                      
    \node[concept,opacity=0] (a) at (0,0){};
    % bot nodes
    \node[concept] (b) at (-0.4,1.5){}; %,fill=yellow!75!orange
    \node[concept] (c) at (0,1){}; %,fill=purple
    \node[concept] (d) at (0.4,1.5){};%,fill=cyan
    % top nodes
    \node[concept,fill=red] (e) at (-1,2){};
    \node[concept,fill=green] (f) at (0,2){};
    \node[concept,fill=blue] (g) at (1,2){};

    % top limit
    \node[concept,opacity=0] (a) at (0,3){};
    \path[edge] (b) edge (e);
    \path[edge] (b) edge (f);
    \path[edge,in=-90,out=150] (c) edge (e);
    \path[edge,in=-90,out=30] (c) edge (g);
    \path[edge] (d) edge (f);
    \path[edge] (d) edge (g);

    \coordinate[label={below left:{\color{yellow!55!orange} \bfseries\tiny y}}]() at (b);
    \coordinate[label={below left:{\color{purple} \bfseries \tiny p}}]() at (c);
    \coordinate[label={below left:{\color{cyan} \bfseries \tiny c}}]() at (d);
  \end{tikzpicture}
  \end{minipage}
  \begin{minipage}{0.2\linewidth}\centering
  \begin{tikzpicture}[concept/.style={fill=black!36,shape=circle,draw=black!80},
                      hidden/.style={opacity=0.5},
                      edge/.style={color=gray},]
    % bottom limit                      
    \node[concept,opacity=0] (a) at (0,0){};

    %cube nodes
    \node[concept,fill=red] (b) at (0.8,2){};
    \node[concept,fill=green] (c) at (1.5,1.15){};
    \node[concept,fill=blue] (d) at (2,2){};

    % top limit
    \node[concept,opacity=0] (a) at (0,3){};

    % cube background pattern
    \draw (1.22,1.8) -- (1.52,1.5);
    \draw (1.32,1.79) -- (1.55,1.56);
    \draw (1.43,1.78) -- (1.58,1.63);
    \draw (1.52,1.79) -- (1.62,1.69);
    \draw (1.6,1.81) -- (1.66,1.75);

    % round cube edges
    \path[in=90,out=-30] (b) edge (c); %,color=yellow!75!orange
    \path[in=210,out=-30] (b) edge (d); %,color=purple
    \path[in=210,out=90] (c) edge (d); %,color=cyan

    \coordinate[label={above left:{\color{purple} \tiny \textbf{p}}}]() at (1.65,2);
    \coordinate[label={above left:{\color{yellow!55!orange} \tiny \textbf{y}}}]() at (1.25,1.3);
    \coordinate[label={above left:{\color{cyan} \tiny \textbf{c}}}]() at (2.1,1.3);
  \end{tikzpicture}
  \end{minipage}
\end{minipage}

%%% Local Variables:
%%% mode: latex
%%% TeX-master: "../paper"
%%% End:
  \caption{The nominal (top) crown (middle) and contranominal (bottom)
    ordinal motif in context (left), concept lattice (middle) and
    geometric drawing style (right) representation. The nominal context
    $\context[N]_{3}$ has an optional $\mathbf{+}$ object and two
    different display styles based on the existence of this
    element. The contranominal ordinal motif has an additional layout,
    namely the tulip layout, for its inner concepts.}
  \label{fig:substructures}
\end{figure}

In \cref{fig:substructures} we show in the first two columns the
nominal ordinal motif on $3+1$ elements, crown ordinal motif on $10$
elements and contranominal ordinal motif on $3$ elements in context
and concept lattice representation. For the contranominal ordinal
motif we depict additionally the \emph{inner} concepts in a different
layout. By inner concepts we refer to the non-top and non-bottom
concepts. We call this layout the \emph{tulip layout}. This layout
results in more readable drawings for join-semilattices, since it is
free of edge crossings.  The last column is discussed in greater
detail in \cref{chapter:topic:sec:geometric}. In short, it employs a
novel geometric technique for drawing lattices.

The nominal ordinal motif is a simple structure that reflects the
incomparability (\cref{fig:substructures}) of the related elements
(topics). A slightly more expressive structure results by adding an
additional (artificial) $+$ object as the meet of all elements of the
motif. This motif can be observed in real world data, for example, it
occurs in \cref{fig:topic-model-concepts} (see concept with the term
\emph{learning}). Equipped with the $+$ version of the ordinal motif,
we will demonstrate a novel geometric drawing technique
\cref{chapter:topic:sec:geometric}.

The crown motif can be identified by its \emph{zig-zag} pattern and
reflects that there is a \emph{round-trip} or a cycle along topics and
documents (see \cref{fig:substructures}, second row).  We were able to
identify many crown ordinal motifs in $\BV_{\text{BS}}$ and
$\BV_{\text{WN}}$. The largest crown ordinal motif in
$\BV_{\text{BS}}$ is a cycle over the topics $($\emph{SVM}, \emph{KM},
\emph{BI}, \emph{O}, \emph{CLass}, \emph{SVM}$)$. In $\BV_{\text{WN}}$
we find many cycles on four elements. For example, one of them is
$($\emph{SemW}, \emph{LKB}, \emph{PR}, \emph{SM}, \emph{SemW}$)$.  The
occurrence of such a motif may reflect a topic based cycle within the
research history of an author. Hence, these cycles constitute
interesting candidates for a temporal topical analysis. In any case,
they are a useful tool to guide readers through an author's research.

Finally, the contranominal ordinal motif can be visually identified
using the tulip layout in the lattice diagrams.  This motif reflects
that there is a unique set of documents for any combination of topics.
Moreover, this type of motif represents a densely explored area within
the topic space. By explored, we refer to the research activities of
the respective entity, e.g., an author.  For example, within
$\BV_{\text{WN}}$ we find that \emph{Social Media}, \emph{Semantic
  Web} and \emph{Learning Knowledge Base}, or \emph{Semantic Web},
\emph{Neural Networks} and \emph{Search Engines} constitute a
contranominal structure.  Furthermore, we find within
$\BV_{\text{BS}}$ the contranominal motifs \emph{Support Vector
  Machines}, \emph{Kernel Methods} and \emph{Optimization}, or
\emph{Support Vector Machines}, \emph{Optimization},
\emph{Classification}.  The involved topics are structurally important
within $\BV_{\text{BS}}$, i.e., within the research of \emph{Bernhard
  Schölkopf}.  Moreover, one may deduce from such structural results
that the occurring topics are highly related within the machine
learning domain. At least they represent a candidate for an important
topic subset.  Beyond the occurrence of ordinal motifs, the absence of
such substructures also carries information.

Particularly important are cases where a motif \emph{almost} occurs,
i.e., adding a few incidences results in a motif.  These may reflect
missing lines of research for future investigations. In the same way,
almost occurring motifs may indicate that important data is missing or
has been filtered in the process. For example while processing the
corpus data with $pq$-core and TITANIC, we may have removed motifs
with low support.

Concluding this analysis, we want to motivate our novel approach
(\cref{chapter:topic:sec:geometric}) by providing a different
geometric interpretation of the motifs. For example, the crown ordinal
motifs reflect a cycle shape of objects (documents) in the topic
space. Hence, one should consider a drawing that reflects this shape
directly. Analogously, the contranominal ordinal motif reflects a
hyperball of documents in the topic space.  Their importance was
addressed in the text above.  Yet, these structures cannot be
(visually) recognized easily in the lattice diagram. Therefor, we
represent them in our novel geometric representation in a unique
shape, i.e., a filled $n$-polygon where $n$ is the dimension of the
hyperball.

\begin{figure}[b!]
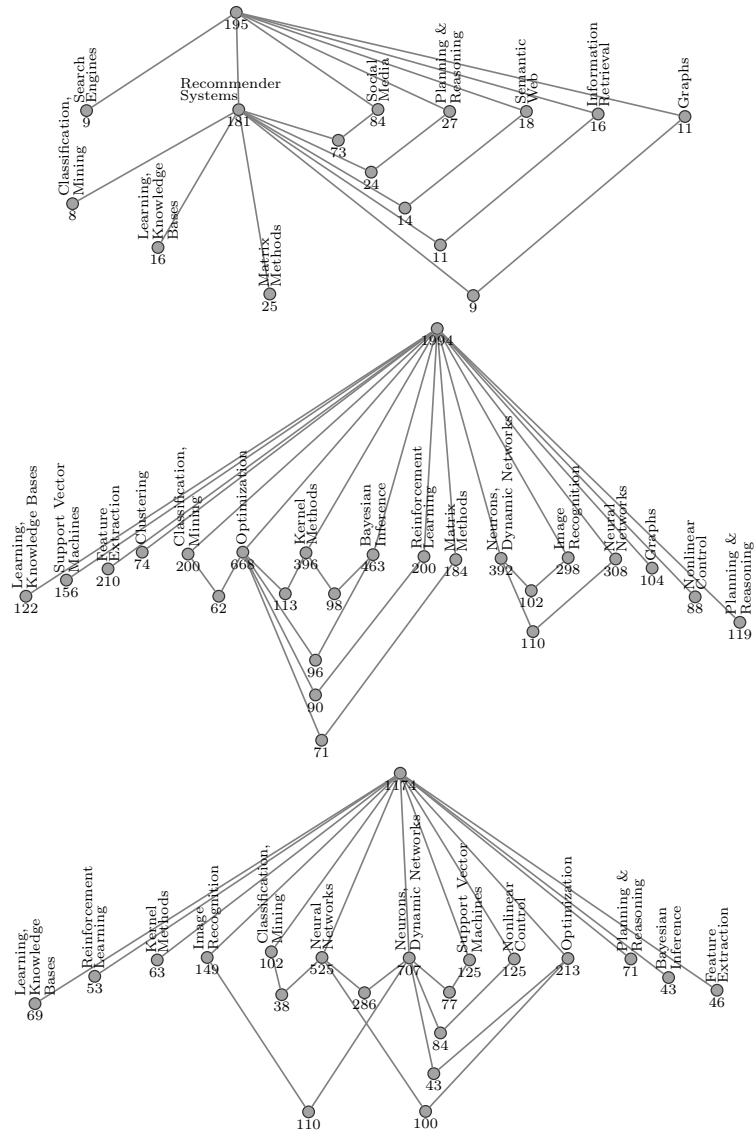

  \centering
  \hspace*{-1cm}\scalebox{0.78}{\input{pics/recsys.tikz}}

  \hspace*{-1cm}\scalebox{0.78}{\input{pics/nips.tikz}}

  \hspace*{-1cm}\scalebox{0.78}{\input{pics/neuralnetworks.tikz}}
  \caption{The concept lattice for  RecSys (top) NeurIPS (middle) and Neural Networks (bot).}
  \label{fig:rs-nips-nn-view}
\end{figure}

\textbf{Ordinal Motifs in Lattices --- Venue Analysis:}\quad %
Analogous to the analysis above, we present an ordinal motif analysis
for the \emph{Recommender Systems} (\emph{RecSys}), \emph{NeurIPS} and \emph{Neural Network}
venues.  We depict their iceberg concept lattices in
\cref{fig:rs-nips-nn-view}.

First, we observe, that the concept hierarchies differ in their
ordinal structure. In particular, we see that the diagram for
\emph{RecSys} is the only one where we attribute annotations on
subconcepts: \emph{Classification, Mining}, \emph{Learning, Knowledge
  Bases} and \emph{Matrix Methods} only occur in concepts where
\emph{Recommender Systems} occurs. This may be interpreted as
\emph{Recommender Systems} dominating the other topics. This relation
constitutes a so far not discussed ordinal motif, called ordinal
ordinal motif [sic] (see \cref{fig:other-motifs}).  We acknowledge
that the dominating role of the \emph{Recommender Systems} topic is
not surprising. Yet, we may point out that this fact was discovered in
an unsupervised fashion, without background information.  A majority
of papers involve this topic, which is not surprising given the title
of the conference. Striking the same chord, we find that the
\emph{Recommender Systems} topic occurs in the most number of
documents. There are numerous nominal ordinal motifs of size two.  The
\emph{Search Engines} topic gives rise to nine nominal motifs without
the $+$ element. We can conclude from this, that the topic
\emph{Search Engines} is isolated in this concept lattice structure.
All other nominal ordinal motifs are in relation to the
\emph{Recommender Systems} topic, e.g., \emph{Social Media} and
\emph{Recommender Systems}, or \emph{Semantic Web} and
\emph{Recommender Systems}. The latter, are in fact nominal $+$ motifs
within the lattice structure, since their meet is present. We observe
no (non-trivial) contranominal or crowns ordinal motifs.  

We want to summarize the novelty of the ordinal approach with respect
to the \emph{RecSys} data. Of course, the identification of
\emph{Recommender Systems} as the most important topic is a simple
question of counting documents and does not require the ordinal
approach. Schäfermeier et al.~\cite{TopicSpaceTrajectories} enabled with their topic
space trajectories (i.e., heatmaps of topic distributions over time)
the identification of co-occurring topics.  Yet, without the
conceptual hierarchy the relation between different topics is
unknown. For example, is \emph{Search Engines} either (1) dominated
by, (2) incomparable to, or (3) coinciding with the topic
\emph{Recommender Systems}?  Given the lattice diagram
(cf. \cref{fig:rs-nips-nn-view}) we can answer this question.  The
analytical building blocks \emph{ordinal motifs} allow for a
structured approach to answering the question above. Moreover, they
enable an automatic extraction of \emph{dominating} topics,
\emph{incomparable} topics, and \emph{incomparable and meet
  coinciding} topics \cite{ordinal-motif-covering}.

For the other two examples entities, i.e., \emph{NeurIPS} and
\emph{Neural Networks}, we observe different results. In short,
different motifs occur, many non-coinciding topics and there are no
dominating topics. Remarkable is the occurrence of contranominal
ordinal motifs. For the \emph{NeurIPS} entity, we find a contranominal
ordinal motif of \emph{Optimization}, \emph{Kernel Methods} and
\emph{Bayesian Inference} and for the \emph{Neural Networks} entity,
we find the contranominal ordinal motif \emph{Neural Network},
\emph{Neurons, Dynamic Networks} and \emph{Optimization}. Both
indicate that there is a strong connection within the respective
topics, i.e., every subset combination of topics occurs. However, all
three contranominal topics do not occur at the same time.

A more global observation is that the conceptual structures of
\emph{NeurIPS} and \emph{Neural Networks} have a larger \emph{width}
compared to \emph{RecSys}. The with of a lattice, or an order relation
in general, is defined as the largest number of elements such that no
two are \emph{comparable}. That is, when no two elements are connected
via a path of only up-ward or down-ward lines. In case of
\emph{RecSys} the width is nine while \emph{Neural Networks} has a
width of thirteen and \emph{NeurIPS} of sixteen.

While \emph{Neural Networks} and \emph{NeurIPS} have similar frequent
topics, their conceptual structure looks quite different. There are
more frequent combinations that involve the topics \emph{Neural
  Network} or \emph{Neural Dynamic Networks} within the view of the
\emph{Neural Networks} entity. From this observation, we can infer
that both entities have a different topic focus.

\subsection{Conceptual Views on Topic Models over Time}\label{sec:topic-view-time}
\begin{figure}
  \centering
  \scalebox{0.95}{\colorlet{mivertexcolor}{black!80}
\colorlet{mivertexcolor}{red}
\colorlet{jivertexcolor}{black!80}
\colorlet{vertexcolor}{black!80}
\colorlet{bordercolor}{black!80}
\colorlet{linecolor}{gray}
% parameter corresponds to the used valuation function and can be addressed by #1
\tikzset{vertexbase/.style 2 args={opacity=#1, semithick, shape=circle, inner sep=2pt, outer sep=0pt, draw=bordercolor},%
  vertex/.style 2 args={vertexbase={#1}{}, fill=vertexcolor!45},
  evertex/.style 2 args={vertexbase={#1}{}, fill=mivertexcolor!45},%
  mivertex/.style 2 args={vertexbase={#1}{}, fill=mivertexcolor!45},%
  jivertex/.style 2 args={vertexbase={#1}{}, fill=jivertexcolor!45},%
  divertex/.style 2 args={vertexbase={#1}{}, top color=mivertexcolor!45, bottom color=jivertexcolor!45},%
  conn/.style={-, thick, color=linecolor},%
  connh/.style={-, opacity=0.2, thick, color=linecolor},%
  connhd/.style={dashed, opacity=0.2, thick, color=linecolor}%
}
\tikzstyle{d} = [text width=1.7cm,align=center]
\tikzstyle{dh} = [opacity=0.2,text width=1.7cm,align=center]
\tikzstyle{h} = [opacity=0.2]
\begin{tikzpicture}[yscale=0.08,xscale=0.18,font=\tiny]
  \begin{scope} %for scaling and the like
    \begin{scope} %draw vertices
      \foreach \nodename/\nodetype/\param/\xpos/\ypos in {%
        0/vertex/0.2/47.99701997459976/-0.9026477947164295,
        1/evertex/1/34.19871313295423/-0.680094458560859,
        2/evertex/1/48.99850998729983/8.889698996128786,
        3/vertex/0.2/35.86786315412103/17.346725770040564,
        4/vertex/0.2/46.43914662151075/23.021835842007675,
        5/vertex/0.2/54.00596005080023/24.46843252701889,
        6/vertex/0.2/44.54744326418838/32.7029059647751,
        7/vertex/1/53.33830004233351/32.81418263285289,
        8/evertex/1/30.3040297502317/42.272699419464736,
        9/vertex/1/70.80873693054599/43.16291276408703,
        10/vertex/0.2/39.4287165326102/44.05312610870932,
        11/vertex/0.2/53.00447003810016/44.832062785253825,
        12/vertex/1/62.017880152400856/46.501212806420625,
        13/evertex/1/24.851473014420172/49.95078951683201,
        14/vertex/0.2/43.32339991533272/50.39589618914315,
        15/evertex/1/32.97466978409859/50.61844952529873,
        16/evertex/1/79.0432103683022/50.61844952529873,
        17/evertex/1/49.55489332768877/50.729726193376514,
        18/vertex/1/68.91703357322362/50.952279529532085,
        19/evertex/1/59.235963450456204/51.39738620184323,
        20/evertex/1/52.25081961885415/69.41111883772174
      } \node[\nodetype={\param}{}] (\nodename) at (\xpos, \ypos) {};
    \end{scope}
    \begin{scope} %draw connections
      \path (12) edge[conn] (19);
      \path (1) edge[conn] (13);
      \path (5) edge[connh] (14);
      \path (17) edge[conn] (20);
      \path (14) edge[connh] (20);
      \path (9) edge[conn] (18);
      \path (19) edge[conn] (20);
      \path (13) edge[conn] (20);
      \path (6) edge[connh] (15);
      \path (10) edge[connh] (15);
      \path (8) edge[conn] (15);
      \path (15) edge[conn] (20);
      \path (11) edge[connh] (19);
      \path (6) edge[connh] (17);
      \path (3) edge[connh] (13);
      \path (11) edge[connh] (17);
      \path (8) edge[conn] (13);
      \path (7) edge[conn] (18);
      \path (18) edge[conn] (20);
      \path (2) edge[conn] (15);
      \path (1) edge[conn] (18);
      \path (4) edge[connh] (19);
      \path (9) edge[conn] (16);
      \path (16) edge[conn] (20);
      \path (3) edge[connh] (14);
      \path (10) edge[connh] (14);
      \path (5) edge[connh] (18);
      \path (0) edge[connh] (16);
      \path (12) edge[conn] (18);
      \path (0) edge[connh] (15);
      \path (2) edge[conn] (18);
      \path (7) edge[conn] (17);
      \path (4) edge[connh] (15);
    \end{scope}
    \begin{scope} %add labels
      \foreach \nodename/\labelpos/\labelopts/\labelcontent in {%
        13/above/d/{Social Media},
        14/above/d/{Learning, Knowledge Bases},
        15/above/d/{Semantic Web},
        16/above/d/{Reinforcement Learning},
        17/above/d/{Neural Networks},
        18/above/d/{Planning \& Reasoning},
        19/above/d/{Search Engines},
        0/below//{},%4
        1/below//{},%4
        2/below//{},%16
        3/below//{},%3
        4/below//{},%14
        5/below//{},%3
        6/below//{},%3
        7/below//{},%3
        8/below//{},%4
        9/below//{},%5
        10/below//{},%6
        11/below//{},%3
        12/below//{},%8
        13/below//{},%2
        14/below//{},%1
        17/below//{},%1
        19/below//{},%2
        0/right/h/{0},
        1/right//{0.167},
        2/right//{0.167},
        3/right/h/{0},
        4/right/h/{0},
        5/right/h/{0},
        6/right/h/{0},
        7/right//{0.167},
        8/right//{0.167},
        9/right//{0.333},
        10/right/h/{0},
        11/right/h/{0},
        12/right//{0.333},
        13/right//{0.167},
        14/right/h/{0},
        15/right//{0.167},
        16/right//{0.333},
        17/right//{0.167},
        18/right//{1},
        19/right//{0.333},
        20/right//{1}
      } \coordinate[label={[\labelopts]\labelpos:{\labelcontent}}](c) at (\nodename);
    \end{scope}
  \end{scope}
\end{tikzpicture}}
  \scalebox{0.95}{\colorlet{mivertexcolor}{black!80}
\colorlet{jivertexcolor}{black!80}
\colorlet{vertexcolor}{black!80}
\colorlet{bordercolor}{black!80}
\colorlet{linecolor}{gray}
% parameter corresponds to the used valuation function and can be addressed by #1
\tikzset{vertexbase/.style 2 args={opacity=#1,semithick, shape=circle, inner sep=2pt, outer sep=0pt, draw=bordercolor},%
  vertex/.style 2 args={vertexbase={#1}{}, fill=vertexcolor!45},%
  mivertex/.style 2 args={vertexbase={#1}{}, fill=mivertexcolor!45},%
  jivertex/.style 2 args={vertexbase={#1}{}, fill=jivertexcolor!45},%
  divertex/.style 2 args={vertexbase={#1}{}, top color=mivertexcolor!45, bottom color=jivertexcolor!45},%
  conn/.style={-, thick, color=linecolor},%
  connh/.style={-, opacity=0.2, thick, color=linecolor}%
}
\tikzstyle{d} = [text width=1.7cm,align=center]
\tikzstyle{dh} = [opacity=0.2,text width=1.7cm,align=center]
\tikzstyle{h} = [opacity=0.2]
\begin{tikzpicture}[yscale=0.08,xscale=0.18,font=\tiny]
  \begin{scope} %for scaling and the like
    \begin{scope} %draw vertices
      \foreach \nodename/\nodetype/\param/\xpos/\ypos in {%
        0/vertex/1/47.99701997459976/-0.9026477947164295,
        1/vertex/1/34.19871313295423/-0.680094458560859,
        2/vertex/1/48.99850998729983/8.889698996128786,
        3/vertex/1/35.86786315412103/17.346725770040564,
        4/vertex/1/46.43914662151075/23.021835842007675,
        5/vertex/1/54.00596005080023/24.46843252701889,
        6/vertex/1/44.54744326418838/32.7029059647751,
        7/vertex/1/53.33830004233351/32.81418263285289,
        8/vertex/0.2/30.3040297502317/42.272699419464736,
        9/vertex/1/70.80873693054599/43.16291276408703,
        10/vertex/1/39.4287165326102/44.05312610870932,
        11/vertex/1/53.00447003810016/44.832062785253825,
        12/vertex/1/62.017880152400856/46.501212806420625,
        13/mivertex/1/24.851473014420172/49.95078951683201,
        14/mivertex/1/43.32339991533272/50.39589618914315,
        15/mivertex/1/32.97466978409859/50.61844952529873,
        16/mivertex/1/79.0432103683022/50.61844952529873,
        17/mivertex/1/49.55489332768877/50.729726193376514,
        18/mivertex/1/68.91703357322362/50.952279529532085,
        19/mivertex/1/59.235963450456204/51.39738620184323,
        20/vertex/1/52.25081961885415/69.41111883772174
      } \node[\nodetype={\param}{}] (\nodename) at (\xpos, \ypos) {};
    \end{scope}
    \begin{scope} %draw connections
      \path (12) edge[conn] (19);
      \path (1) edge[conn] (13);
      \path (5) edge[conn] (14);
      \path (17) edge[conn] (20);
      \path (14) edge[conn] (20);
      \path (9) edge[conn] (18);
      \path (19) edge[conn] (20);
      \path (13) edge[conn] (20);
      \path (6) edge[conn] (15);
      \path (10) edge[conn] (15);
      \path (8) edge[connh] (15);
      \path (15) edge[conn] (20);
      \path (11) edge[conn] (19);
      \path (6) edge[conn] (17);
      \path (3) edge[conn] (13);
      \path (11) edge[conn] (17);
      \path (8) edge[connh] (13);
      \path (7) edge[conn] (18);
      \path (18) edge[conn] (20);
      \path (2) edge[conn] (15);
      \path (1) edge[conn] (18);
      \path (4) edge[conn] (19);
      \path (9) edge[conn] (16);
      \path (16) edge[conn] (20);
      \path (3) edge[conn] (14);
      \path (10) edge[conn] (14);
      \path (5) edge[conn] (18);
      \path (0) edge[conn] (16);
      \path (12) edge[conn] (18);
      \path (0) edge[conn] (15);
      \path (2) edge[conn] (18);
      \path (7) edge[conn] (17);
      \path (4) edge[conn] (15);
    \end{scope}
    \begin{scope} %add labels
      \foreach \nodename/\labelpos/\labelopts/\labelcontent in {%
        13/above/d/{Social Media},
        14/above/d/{Learning, Knowledge Bases},
        15/above/d/{Semantic Web},
        16/above/d/{Reinforcement Learning},
        17/above/d/{Neural Networks},
        18/above/d/{Planning \& Reasoning},
        19/above/d/{Search Engines},
        0/below//{},%4
        1/below//{},%4
        2/below//{},%16
        3/below//{},%3
        4/below//{},%14
        5/below//{},%3
        6/below//{},%3
        7/below//{},%3
        8/below//{},%4
        9/below//{},%5
        10/below//{},%6
        11/below//{},%3
        12/below//{},%8
        13/below//{},%2
        14/below//{},%1
        17/below//{},%1
        19/below//{},%2
        0/right//{0.065},
        1/right//{0.043},
        2/right//{0.326},
        3/right//{0.043},
        4/right//{0.130},
        5/right//{0.065},
        6/right//{0.065},
        7/right//{0.043},
        8/right/h/{0},
        9/right//{0.043},
        10/right//{0.065},
        11/right//{0.065},
        12/right//{0.109},
        13/right//{0.087},
        14/right//{0.152},
        15/right//{0.630},
        16/right//{0.087},
        17/right//{0.196},
        18/right//{0.587},
        19/right//{0.304},
        20/right//{1}
      } \coordinate[label={[\labelopts]\labelpos:{\labelcontent}}](c) at (\nodename);
    \end{scope}
  \end{scope}
\end{tikzpicture}}
  \scalebox{0.95}{\colorlet{mivertexcolor}{black!80}
\colorlet{jivertexcolor}{black!80}
\colorlet{vertexcolor}{black!80}
\colorlet{bordercolor}{black!80}
\colorlet{linecolor}{gray}
% parameter corresponds to the used valuation function and can be addressed by #1
\tikzset{vertexbase/.style 2 args={opacity=#1, semithick, shape=circle, inner sep=2pt, outer sep=0pt, draw=bordercolor},%
  vertex/.style 2 args={vertexbase={#1}{}, fill=vertexcolor!45},%
  mivertex/.style 2 args={vertexbase={#1}{}, fill=mivertexcolor!45},%
  jivertex/.style 2 args={vertexbase={#1}{}, fill=jivertexcolor!45},%
  divertex/.style 2 args={vertexbase={#1}{}, top color=mivertexcolor!45, bottom color=jivertexcolor!45},%
  conn/.style={-, thick, color=linecolor},%
  connh/.style={-, opacity=0.2, thick, color=linecolor}%
}
\tikzstyle{d} = [text width=1.7cm,align=center]
\tikzstyle{dh} = [opacity=0.2,text width=1.7cm,align=center]
\tikzstyle{h} = [opacity=0.2]
\begin{tikzpicture}[yscale=0.08,xscale=0.18,font=\tiny]
  \begin{scope} %for scaling and the like
    \begin{scope} %draw vertices
      \foreach \nodename/\nodetype/\param/\xpos/\ypos in {%
        0/vertex/1/47.99701997459976/-0.9026477947164295,
        1/vertex/1/34.19871313295423/-0.680094458560859,
        2/vertex/0.2/48.99850998729983/8.889698996128786,
        3/vertex/1/35.86786315412103/17.346725770040564,
        4/vertex/1/46.43914662151075/23.021835842007675,
        5/vertex/0.2/54.00596005080023/24.46843252701889,
        6/vertex/0.2/44.54744326418838/32.7029059647751,
        7/vertex/0.2/53.33830004233351/32.81418263285289,
        8/vertex/1/30.3040297502317/42.272699419464736,
        9/vertex/1/70.80873693054599/43.16291276408703,
        10/vertex/1/39.4287165326102/44.05312610870932,
        11/vertex/0.2/53.00447003810016/44.832062785253825,
        12/vertex/1/62.017880152400856/46.501212806420625,
        13/mivertex/1/24.851473014420172/49.95078951683201,
        14/mivertex/1/43.32339991533272/50.39589618914315,
        15/mivertex/1/32.97466978409859/50.61844952529873,
        16/mivertex/1/79.0432103683022/50.61844952529873,
        17/mivertex/0.2/49.55489332768877/50.729726193376514,
        18/mivertex/1/68.91703357322362/50.952279529532085,
        19/mivertex/1/59.235963450456204/51.39738620184323,
        20/vertex/1/52.25081961885415/69.41111883772174
      } \node[\nodetype={\param}{}] (\nodename) at (\xpos, \ypos) {};
    \end{scope}
    \begin{scope} %draw connections
      \path (12) edge[conn] (19);
      \path (1) edge[conn] (13);
      \path (5) edge[connh] (14);
      \path (17) edge[connh] (20);
      \path (14) edge[conn] (20);
      \path (9) edge[conn] (18);
      \path (19) edge[conn] (20);
      \path (13) edge[conn] (20);
      \path (6) edge[connh] (15);
      \path (10) edge[conn] (15);
      \path (8) edge[conn] (15);
      \path (15) edge[conn] (20);
      \path (11) edge[connh] (19);
      \path (6) edge[connh] (17);
      \path (3) edge[conn] (13);
      \path (11) edge[connh] (17);
      \path (8) edge[conn] (13);
      \path (7) edge[connh] (18);
      \path (18) edge[conn] (20);
      \path (2) edge[connh] (15);
      \path (1) edge[conn] (18);
      \path (4) edge[conn] (19);
      \path (9) edge[conn] (16);
      \path (16) edge[conn] (20);
      \path (3) edge[conn] (14);
      \path (10) edge[conn] (14);
      \path (5) edge[connh] (18);
      \path (0) edge[conn] (16);
      \path (12) edge[conn] (18);
      \path (0) edge[conn] (15);
      \path (2) edge[connh] (18);
      \path (7) edge[connh] (17);
      \path (4) edge[conn] (15);
    \end{scope}
    \begin{scope} %add labels
      \foreach \nodename/\labelpos/\labelopts/\labelcontent in {%
        13/above/d/{Social Media},
        14/above/d/{Learning, Knowledge Bases},
        15/above/d/{Semantic Web},
        16/above/d/{Reinforcement Learning},
        17/above/dh/{Neural Networks},
        18/above/d/{Planning \& Reasoning},
        19/above/d/{Search Engines},
        0/below//{},%4
        1/below//{},%4
        2/below//{},%16
        3/below//{},%3
        4/below//{},%14
        5/below//{},%3
        6/below//{},%3
        7/below//{},%3
        8/below//{},%4
        9/below//{},%5
        10/below//{},%6
        11/below//{},%3
        12/below//{},%8
        13/below//{},%2
        14/below//{},%1
        17/below//{},%1
        19/below//{},%2
        0/right//{0.048},
        1/right//{0.048},
        2/right/h/{0},
        3/right//{0.048},
        4/right//{0.381},
        5/right/h/{0},
        6/right/h/{0},
        7/right/h/{0},
        8/right//{0.143},
        9/right//{0.048},
        10/right//{0.143},
        11/right/h/{0},
        12/right//{0.048},
        13/right//{0.286},
        14/right//{0.238},
        15/right//{0.714},
        16/right//{0.095},
        17/right/h/{0},
        18/right//{0.143},
        19/right//{0.524},
        20/right//{1}
      } \coordinate[label={[\labelopts]\labelpos:{\labelcontent}}](c) at (\nodename);
    \end{scope}
  \end{scope}
\end{tikzpicture}}
  \caption{The concept lattice for Wolgang Nejdls. The support of
    each concept for the years 1987-1999, 2000-2008 and 2009-2020 is annotated
    next to each concept. Attribute sets that are not closed in a
    given time interval are highlighted in red.  }
  \label{fig:nejdl-temporal-view}
\end{figure}

A particular feature of the just introduced methods is that they
enable an investigations over time. We demonstrate this on two
examples.  First, we consider the conceptual structure of Wolfgang
Nejdl, as depicted in \cref{fig:nejdl-temporal-view}.  In this figure,
we show the diagram for three different time periods. 
These periods were chosen based on the following observation within
the heatmap in \cref{fig:topic-model1}. We empirically identified
three main periods in \emph{Wolgang Nejdls} research history. The
first is from 1987-1999 where he researched mainly on \emph{Planning
  \& Reasoning}. Second, the period from 2000 to 2008 which seems to
be a transition phase where both the \emph{Planning \& Reasoning} and
\emph{Semantic Web} topics are present. Lastly, there is the period
starting with 2009 where he focused primarily on \emph{Semantic Web}.

For each period, we annotated at concepts their support value, i.e.,
the relative number of research articles for the corresponding topics.
We highlighted the supported concepts (or toned down the non-supported
concepts) to make the resemblance to the original structure more
clear.  It is important to note that we encountered the case that a
set of topics is not closed for the period 1987-1999. Nonetheless, we
stick to the common conceptual representation, but highlighted
non-closed topic sets in red \cite{smeasure-error}. 
% cref chapter:error

% 1
Based on the support values, we see that \emph{Planning \& Reasoning}
is the most important topic in the first period. Even more, as we can
infer from the non-red nodes, all topics coincide with \emph{Planning
  \& Reasoning}. The second most frequent topics are \emph{Search
  Engines} and \emph{Reinforcement Learning}. Another observation we
can draw is that a majority of the topic combinations are not
supported or closed at this stage.
% 2
In the second time period, we see that almost all topic combinations
are supported. The exception is the combination of \emph{Social Media}
and \emph{Semantic Web}. The \emph{Planning \& Reasoning} topic is
less supported. The \emph{Semantic Web} topic has the highest support
value.
% 3
In the last period the activity on \emph{Planning \& Reasoning}
declines again. Five of eight topic combinations involve the
\emph{Semantic Web} topic.  Compared to the second diagram, fewer
topic combinations (eight compared to twelve) are supported. Overall
we see a shift in activity from concepts depicted on right to concepts
on the left.

\begin{figure}[H]
  \centering
  \colorlet{mivertexcolor}{black!80}
\colorlet{mivertexcolor}{red}
\colorlet{jivertexcolor}{black!80}
\colorlet{vertexcolor}{black!80}
\colorlet{bordercolor}{black!80}
\colorlet{linecolor}{gray}
% parameter corresponds to the used valuation function and can be addressed by #1
\tikzset{vertexbase/.style 2 args={semithick, shape=circle, inner sep=2pt, outer sep=0pt, draw=bordercolor},%
  vertex/.style 2 args={vertexbase={#1}{}, fill=vertexcolor!45},%
  evertex/.style 2 args={vertexbase={#1}{}, fill=mivertexcolor!45},%
  mivertex/.style 2 args={vertexbase={#1}{}, fill=mivertexcolor!45},%
  jivertex/.style 2 args={vertexbase={#1}{}, fill=jivertexcolor!45},%
  divertex/.style 2 args={vertexbase={#1}{}, top color=mivertexcolor!45, bottom color=jivertexcolor!45},%
  conn/.style={-, thick, color=linecolor},%
  connh/.style={-, opacity=0.2, thick, color=linecolor}%
}
\tikzstyle{d} = [rotate=90]
\tikzstyle{dh} = [opacity=0.2,text width=1.7cm,align=center]
\tikzstyle{h} = [opacity=0.2]
\begin{tikzpicture}[xscale=0.25,yscale=0.15,font=\tiny]
  \begin{scope} %for scaling and the like
    \begin{scope} %draw vertices
      \foreach \nodename/\nodetype/\param/\xpos/\ypos in {%
        0/vertex/1/27.876551729887613/29.34157140064901,
        1/vertex/1/14.003849855630406/29.513955726660257,
        2/vertex/1/6.381135707410955/34.807507218479316,
        3/vertex/1/25.634032130917323/35.09162165441898,
        4/vertex/1/23.219011024333934/39.289158339671054,
        5/vertex/1/0.5582290664099929/39.78344562078924,
        6/vertex/1/20.918990922825945/43.37169401984774,
        7/vertex/1/18.67647132385566/46.99422567972282,
        8/vertex/1/42.309177866850234/49.6967492989947,
        9/vertex/1/36.38662610546717/49.984251811683194,
        10/vertex/1/26.266537658832018/50.2717543243717,
        11/vertex/1/31.499083389762692/50.2717543243717,
        12/vertex/1/1.5110683349374199/50.37054860442736,
        13/vertex/1/11.88642925890278/50.47641963426374,
        14/vertex/1/21.378994943127548/50.5017563345225,
        15/vertex/1/11.718910516793994/61.54185282176084
      } \node[\nodetype={\param}{}] (\nodename) at (\xpos, \ypos) {};
    \end{scope}
    \begin{scope} %draw connections
      \path (12) edge[conn] (15);
      \path (5) edge[conn] (13);
      \path (1) edge[conn] (13);
      \path (8) edge[conn] (15);
      \path (3) edge[conn] (9);
      \path (4) edge[conn] (13);
      \path (2) edge[conn] (13);
      \path (7) edge[conn] (14);
      \path (6) edge[conn] (10);
      \path (10) edge[conn] (15);
      \path (7) edge[conn] (13);
      \path (14) edge[conn] (15);
      \path (0) edge[conn] (8);
      \path (6) edge[conn] (13);
      \path (13) edge[conn] (15);
      \path (3) edge[conn] (13);
      \path (11) edge[conn] (15);
      \path (0) edge[conn] (13);
      \path (4) edge[conn] (11);
      \path (9) edge[conn] (15);
    \end{scope}
    \begin{scope} %add labels
      \foreach \nodename/\labelpos/\labelopts/\labelcontent in {%
        1/right/d/{\parbox{2cm}{\baselineskip=0pt Matrix\\ Methods}},
        2/right/d/{\parbox{2cm}{\baselineskip=0pt Learning,\\ Knowledge Bases}},
        5/right/d/{\parbox{2cm}{\baselineskip=0pt Classification,\\ Mining}},
        8/right/d/{Graphs},
        9/right/d/{\parbox{2cm}{\baselineskip=0pt Information\\ Retrieval}},
        10/right/d/{\parbox{2cm}{\baselineskip=0pt Planning \&\\ Reasoning}},
        11/right/d/{\parbox{2cm}{\baselineskip=0pt Semantic\\ Web}},
        12/right/d/{\parbox{2cm}{\baselineskip=0pt Search\\ Engines}},
        13/above//{\parbox{2cm}{\baselineskip=0pt Recommender\\ Systems}},
        14/right/d/{\parbox{2cm}{\baselineskip=0pt Social\\ Media}},
        0/below//{},%9
        1/below//{},%25
        2/below//{},%16
        3/below//{},%11
        4/below//{},%14
        5/below//{},%8
        6/below//{},%24
        7/below//{},%73
        8/below//{},%2
        9/below//{},%5
        10/below//{},%3
        11/below//{},%4
        12/below//{},%9
        13/below//{},%5
        14/below//{},%11
        0/right//{0.053},
        1/right//{0.124},
        2/right//{0.035},
        3/right//{0.062},
        4/right//{0.071},
        5/right//{0.053},
        6/right//{0.115},
        7/right//{0.389},
        8/right//{0.062},
        9/right//{0.097},
        10/right//{0.124},
        11/right//{0.106},
        12/right//{0.053},
        13/right//{0.903},
        14/right//{0.478},
        15/right//{1}
      } \coordinate[label={[\labelopts]\labelpos:{\labelcontent}}](c) at (\nodename);
    \end{scope}
  \end{scope}
\end{tikzpicture}
  \colorlet{mivertexcolor}{black!80}
\colorlet{mivertexcolor}{red}
\colorlet{jivertexcolor}{black!80}
\colorlet{vertexcolor}{black!80}
\colorlet{bordercolor}{black!80}
\colorlet{linecolor}{gray}
% parameter corresponds to the used valuation function and can be addressed by #1
\tikzset{vertexbase/.style 2 args={semithick, shape=circle, inner sep=2pt, outer sep=0pt, draw=bordercolor},%
  vertex/.style 2 args={vertexbase={#1}{}, fill=vertexcolor!45},%
  evertex/.style 2 args={vertexbase={#1}{}, fill=mivertexcolor!45},%
  mivertex/.style 2 args={vertexbase={#1}{}, fill=mivertexcolor!45},%
  jivertex/.style 2 args={vertexbase={#1}{}, fill=jivertexcolor!45},%
  divertex/.style 2 args={vertexbase={#1}{}, top color=mivertexcolor!45, bottom color=jivertexcolor!45},%
  conn/.style={-, thick, color=linecolor}%
}
\tikzstyle{d} = [rotate=90]
\begin{tikzpicture}[xscale=0.25,yscale=0.15,font=\tiny]
  \begin{scope} %for scaling and the like
    \begin{scope} %draw vertices
      \foreach \nodename/\nodetype/\param/\xpos/\ypos in {%
        0/vertex/1/27.876551729887613/29.34157140064901,
        1/vertex/1/14.003849855630406/29.513955726660257,
        2/vertex/1/6.381135707410955/34.807507218479316,
        3/vertex/1/25.634032130917323/35.09162165441898,
        4/vertex/1/23.219011024333934/39.289158339671054,
        5/vertex/1/0.5582290664099929/39.78344562078924,
        6/vertex/1/20.918990922825945/43.37169401984774,
        7/vertex/1/18.67647132385566/46.99422567972282,
        8/vertex/1/42.309177866850234/49.6967492989947,
        9/vertex/1/36.38662610546717/49.984251811683194,
        10/vertex/1/26.266537658832018/50.2717543243717,
        11/evertex/1/31.499083389762692/50.2717543243717,
        12/vertex/1/1.5110683349374199/50.37054860442736,
        13/vertex/1/11.88642925890278/50.47641963426374,
        14/vertex/1/21.378994943127548/50.5017563345225,
        15/vertex/1/11.718910516793994/61.54185282176084
      } \node[\nodetype={\param}{}] (\nodename) at (\xpos, \ypos) {};
    \end{scope}
    \begin{scope} %draw connections
      \path (12) edge[conn] (15);
      \path (5) edge[conn] (13);
      \path (1) edge[conn] (13);
      \path (8) edge[conn] (15);
      \path (3) edge[conn] (9);
      \path (4) edge[conn] (13);
      \path (2) edge[conn] (13);
      \path (7) edge[conn] (14);
      \path (6) edge[conn] (10);
      \path (10) edge[conn] (15);
      \path (7) edge[conn] (13);
      \path (14) edge[conn] (15);
      \path (0) edge[conn] (8);
      \path (6) edge[conn] (13);
      \path (13) edge[conn] (15);
      \path (3) edge[conn] (13);
      \path (11) edge[conn] (15);
      \path (0) edge[conn] (13);
      \path (4) edge[conn] (11);
      \path (9) edge[conn] (15);
    \end{scope}
    \begin{scope} %add labels
      \foreach \nodename/\labelpos/\labelopts/\labelcontent in {%
        1/right/d/{\parbox{2cm}{\baselineskip=0pt Matrix\\ Methods}},
        2/right/d/{\parbox{2cm}{\baselineskip=0pt Learning,\\ Knowledge Bases}},
        5/right/d/{\parbox{2cm}{\baselineskip=0pt Classification,\\ Mining}},
        8/right/d/{Graphs},
        9/right/d/{\parbox{2cm}{\baselineskip=0pt Information\\ Retrieval}},
        10/right/d/{\parbox{2cm}{\baselineskip=0pt Planning \&\\ Reasoning}},
        11/right/d/{\parbox{2cm}{\baselineskip=0pt Semantic\\ Web}},
        12/right/d/{\parbox{2cm}{\baselineskip=0pt Search\\ Engines}},
        13/above//{\parbox{2cm}{\baselineskip=0pt Recommender\\ Systems}},
        14/right/d/{\parbox{2cm}{\baselineskip=0pt Social\\ Media}},
        0/below//{},%9
        1/below//{},%25
        2/below//{},%16
        3/below//{},%11
        4/below//{},%14
        5/below//{},%8
        6/below//{},%24
        7/below//{},%73
        8/below//{},%2
        9/below//{},%5
        10/below//{},%3
        11/below//{},%4
        12/below//{},%9
        13/below//{},%5
        14/below//{},%11
        0/right//{0.037},
        1/right//{0.134},
        2/right//{0.146},
        3/right//{0.049},
        4/right//{0.073},
        5/right//{0.034},
        6/right//{0.134},
        7/right//{0.354},
        8/right//{0.049},
        9/right//{0.061},
        10/right//{0.159},
        11/right//{0.073},
        12/right//{0.037},
        13/right//{0.963},
        14/right//{0.366},
        15/right//{1}
      } \coordinate[label={[\labelopts]\labelpos:{\labelcontent}}](c) at (\nodename);
    \end{scope}
  \end{scope}
\end{tikzpicture}
  \caption{The concept lattice for the RecSys venue. The support of
    each concept for the years 1987-2014 and 2015-2020 is annotated
    next to each concept. Attribute sets that are not closed in a
    given time interval are highlighted in red.}
  \label{fig:recsys-temporal-view}
\end{figure}
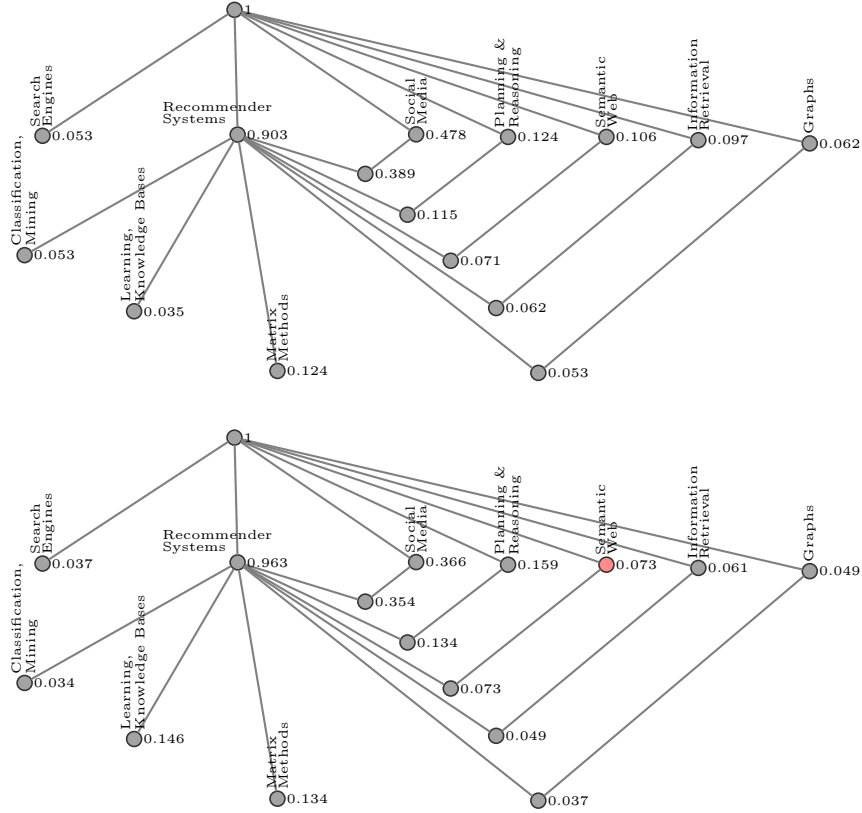

We approach \emph{RecSys} in the same way (see
\cref{fig:recsys-temporal-view}).  Here we chose the two time periods
1987-2014 and 2015-2020. In contrast to \emph{Wolfgang Nejdl}, we do
not observe notable conceptual differences over time. One may take
this for evidence that \emph{RecSys} has a stable focus.

\subsection{Association Rules in Conceptual Topic Views}
The introduced conceptual structures allow for the extraction of
implicational rules between topics.  These are rules of the form
$A\limplies B$ where $A,B\subseteq \Topics$ and state that all
documents $\documnt \in \Documents$ that are in relation with $A$,
i.e., $(\documnt,\topic)\in \IDT$ for all $\topic \in A$, are also in
relation with $B$.
The validity of a rule with respect to our data set
$\IDT$ can be quantified by two measures, namely, the \emph{support} and
\emph{confidence} of $A\limplies B$. The support is the
relative number of documents $\documnt\in \Documents$ that are in
relation to $A$, i.e., the support of $A$ in $\Documents$:
$\supp(A\limplies B)\coloneqq \supp(A)\coloneqq
\nicefrac{\abs{\{\documnt\in \Documents\mid \forall \topic\in
    A:(\documnt,\topic)\in \IDT\}}}{\abs{\Documents}}$. The confidence
is the relative number of documents for which the rule holds,
i.e.,
$\conf(A\limplies B)\coloneqq \nicefrac{\supp(A\cup B)}{\supp{A}}$.

Due to the large number of valid rules in a data set, a standard
approach is to compute a minimal number of rules from which all other
rules can be deduced. Such a set is called a \emph{basis}.  The
Luxenburg basis~\cite{luxenburger-basis} computes a basis that
satisfies a minimum support and confidence value.  That is, given a
minimum support $\eta$ and minimum confidence $\gamma$ all rules that
can be deduced from the Luxenburger basis have $\eta$ and $\gamma$.
For computing the Luxenburger basis for the entities of $\IDT$ (see
\cref{chapter:topic:sec:reduction}) we chose a minimum support of
three percent. This is the same parameter as for the TITANIC algorithm
in \cref{chapter:topic:sec:reduction}. Hence, the computed rules are
reflected by the iceberg concept lattices diagrams
(cf. \cref{fig:schoelkopf-nejdl-view,fig:rs-nips-nn-view}).  For
$\gamma$ we chose fifty percent in order to find meaningful rules.

\begin{table}[t]
  \centering
  \caption{The luxenburg basis for $\IDT$ and entities from
    \cref{chapter:topic:sec:reduction} for a minimum support three percent and
    minimum confidence of fifty percent.}
  \small
  \setlength\tabcolsep{2pt}
  \begin{tabular}{|rcl|l|l|}
    \hline
    Luxenburg Basis&&& Support & Confidence\\ \hline
    \hline
    \multicolumn{5}{|c|}{Bernhard Schölkopf}\\\hline
    Support Vector Machines &$\limplies$& Kernel Methods& 0.48& 0.50\\
    Kernel Methods& $\limplies$ &Support Vector Machines & 0.40 & 0.59\\ \hline    
    \hline
    \multicolumn{5}{|c|}{Wolfgang Nejdl}\\\hline 
    $\emptyset$ &$\limplies$& Semantic Web & 1.00 & 0.61\\
    Reinforcement Learning &$\limplies$& Planning \& Reasoning & 0.10 &  0.62\\
    Search Engines &$\limplies$& Semantic Web & 0.36 & 0.51\\
    Reinforcement Learning &$\limplies$&  Semantic Web & 0.10 & 0.50\\
    Learning Knowledge Bases &$\limplies$& Semantic Web & 0.16 &  0.50\\\hline
    \hline
    \multicolumn{5}{|c|}{RecSys}\\\hline 
 $\emptyset$ &$\limplies$& Recommender Systems & 1.00 & 0.92\\
 Information Retrieval &$\limplies$& Recommender Systems &  0.08 &  0.68\\
 Planning \& Reasoning &$\limplies$& Recommender Systems & 0.13 & 0.88\\
 Semantic Web &$\limplies$& Recommender Systems & 0.09 & 0.77\\
 Social Media &$\limplies$& Recommender Systems & 0.43 & 0.86\\
 Graphs &$\limplies$& Recommender Systems & 0.05 & 0.81\\\hline
    \hline
    \multicolumn{5}{|c|}{Neural Networks}\\\hline 
 $\emptyset$ &$\limplies$& Neurons Dynamic Networks & 1.00 & 0.60\\
 Image Recognition &$\limplies$& Neurons Dynamic Networks & 0.12 & 0.73\\
 Support Vector Machines &$\limplies$& Neurons Dynamic Networks  & 0.10 & 0.61\\
 Neural Networks &$\limplies$& Neurons Dynamic Networks  & 0.44 & 0.54\\
 Nonlinear Control &$\limplies$& Neurons Dynamic Networks  & 0.10 & 0.67\\\hline
    \hline
    \multicolumn{5}{|c|}{NeurIPS}\\ \hline  
%    \multicolumn{5}{|c|}{}\\  
   \hline
  \end{tabular}
  \label{tab:topic-rules}
\end{table}

The resulting bases are depicted in \cref{tab:topic-rules}. In the
following, we discuss the results for the considered entities.  For
\emph{Bernhard Schölkopf} the rules identify a strong inter-dependence
between \emph{Support Vector Machines} and \emph{Kernel Methods}.
This is consistent to our findings in \cref{sec:bv_analysis}.  For
\emph{Wolfgang Nejdl}, we first note the overall importance of the
\emph{Semantice Web} topic. Four out of five rules have
\emph{Semantice Web} in their head. Yet, this topic never occurs in
the body of a rule. This is in contrast to the topics of
\emph{Bernhard Schölkopf}. For \emph{RecSys}, the found rules confirm
our results in \cref{sec:bv_analysis}, since all rules have the topic
\emph{Recommender Systems} in their head. The same applies to
\emph{Neural Networks}, where all rules have \emph{Neurons Dynamic
  Networks} in their head. 
\emph{NeurIPS} on the other hand does not have any rules in the given
basis for the given parameters. This may indicate that the NeurIPS is
a topic diverse venue within the field of ML.

We may note that the support values of the given topic combinations
can also be read directly from the concept diagrams (see
\cref{chapter:topic:sec:view-interpretation}). However, the computed
rules allow for a more comprehensive representation of the most
confident topic dependencies.

\subsection{The Conceptual Term-Topic Structure}\label{chapter:topic:sec:term-topic-view}

\begin{figure}[H]
  \centering
  \includegraphics[trim=120 0 120 0,clip,width=0.6\linewidth]{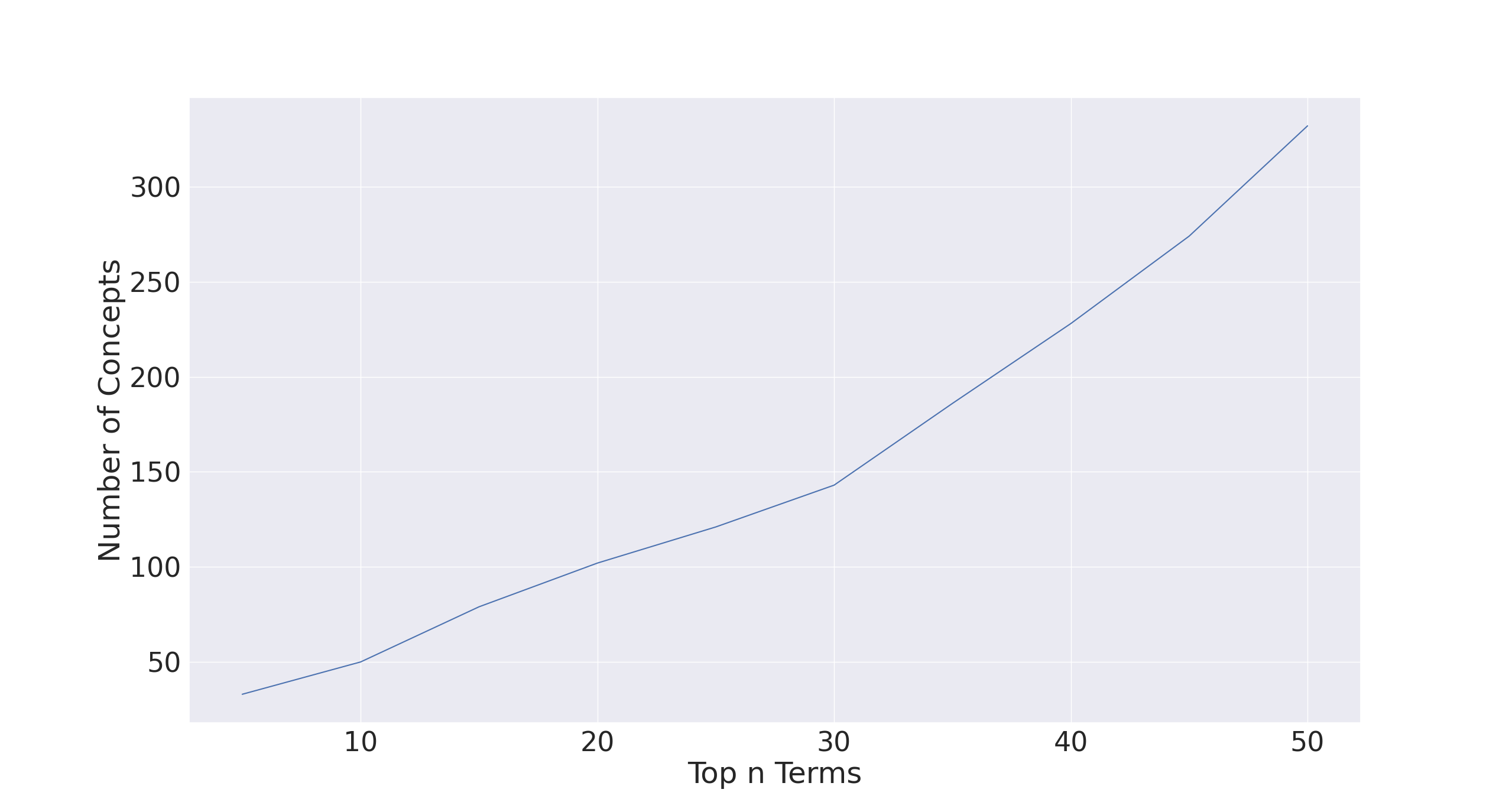}
  \caption{The number of formal concepts of $\ITT$ depending on the top
    terms parameter $n$.}
  \label{fig:top-n-term-concepts}
\end{figure}

The topic-term relation $\ITT$ (see \cref{fig:topic-term-matrix})
entails important information on the \SSHTM topic model. It allows us
to explain the topics of \SSHTM via terms $\term \in \Terms$. As
discussed in \cref{chapter:topic:sec:relation}, we derive an incidence
structure $\ITT_{n}$ from $\ITT$. This parameters is the number of
top-$n$ terms per topic.  Our choice for the $n\in \mathbb{N}$ depends
on the corresponding number of formal concepts, as depicted in
\cref{fig:top-n-term-concepts}. From the plot, we infer that the
parameter of $n=10$ (see $\ITT_{10}$ \cref{fig:top-ten-terms}) is
reasonable, as it results in about fifty formal concepts. This
parameter choice is common in the literature
\cite{TopicSpaceTrajectories,TM+FCA,coherencemodel}, independently of
our requirement on a low number of concepts.

\begin{figure}
  \centering
  \includegraphics[height=0.9\textheight]{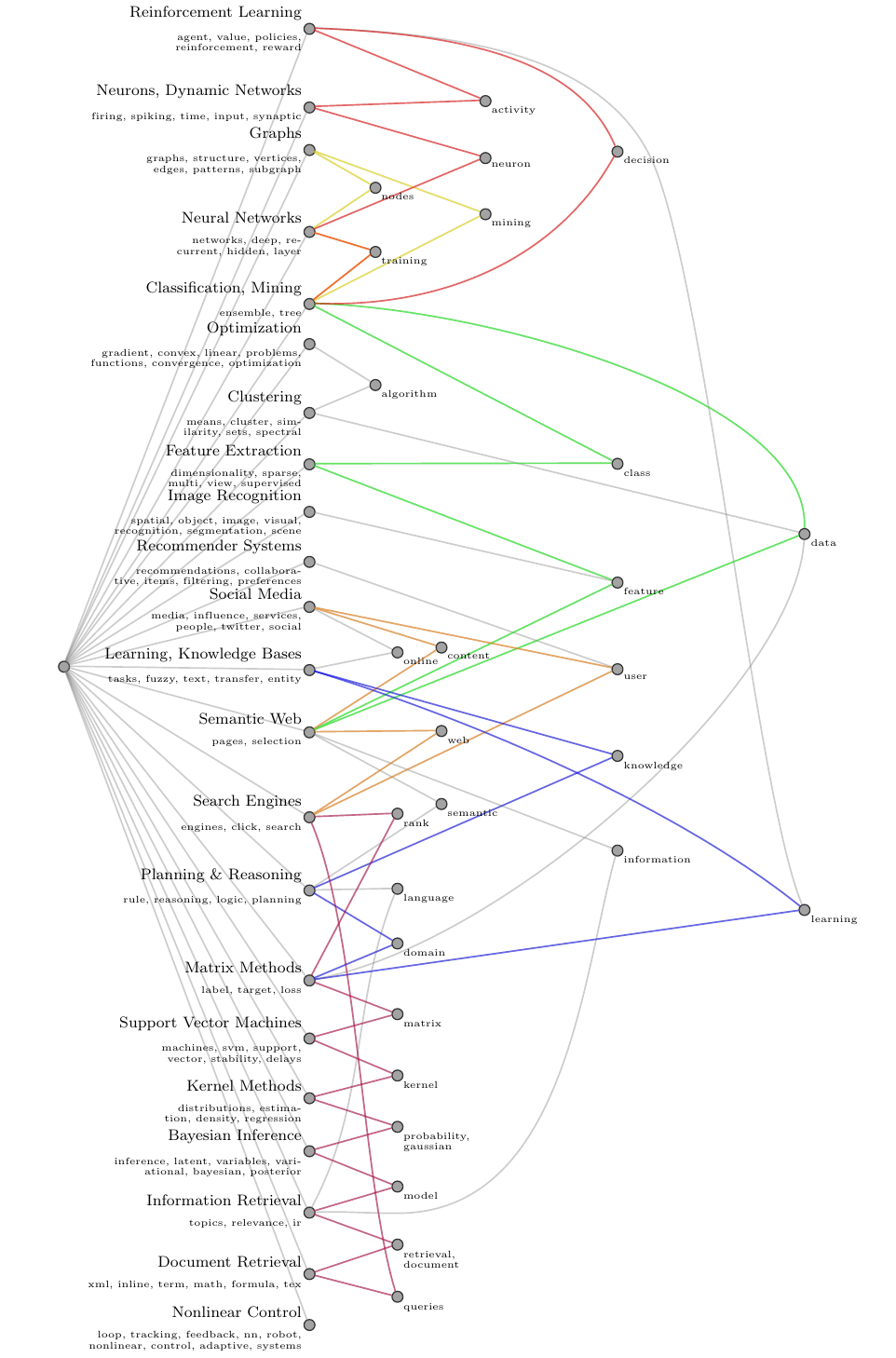}
  \caption{The concept lattice of the term-topic relation of the
    \SSHTM topic model.}
  % there are more crowns: On the left with information
  % A very large one combining the red with the top part of blue and green, blue green and orange, as well as all colored in a zig zag fashion
  % cube information
  \label{fig:topic-model-concepts}
\end{figure}

We depict the concept lattice of $\ITT_{10}$ in
\cref{fig:topic-model-concepts}. We omitted to present the bottom
concept, since it was not supported.  An advantage of this structure
is that we can explain found topic dependencies in terms of their
shared terms.  For example, the topics \emph{Kernel Methods} and
\emph{Support Vector Machines} are connected via the term
\emph{kernels}.  Analogously, the topics \emph{Neural Networks} and
\emph{Graphs} are connected via the term \emph{nodes}. 

We highlighted several contranominal and crown ordinal motifs using
different colors. For example, the topics \emph{Search Engines},
\emph{Semantic Web} and \emph{Social Media} are of contranominal
structure. For this, the terms \emph{web}, \emph{user} and
\emph{content} are responsible (highlighted in orange).
Another example is the set of topics \emph{Planning \& Reasoning},
\emph{Matrix Methods} and \emph{Leaning Knowledge Bases}, which are
also of contranominal structure. For this motif, the terms
\emph{domain}, \emph{learning} and \emph{knowledge} are responsible
(highlighted in blue). 
Hence, these terms are pair-wise differently used in the \SSHTM topic
model, yet they are very similar. One may deduce from this observation
that this is also true for the research corpus $\Documents$.
An example crown motif is given by the sub-structure highlighted in
red (right). This motif spans from the topic \emph{Classification} 
over the topics \emph{Neural Networks}, \emph{Neurons and Dynamic Networks} to
\emph{Reinforcement Learning} and back to the \emph{Classification}
topic. A larger crown is depicted in purple on the left in the
diagram. 

The proposed method allows for meaningful and structural
investigations of the \SSHTM topic model. This distinguishes our
method from other approaches, such as those presented in
\cref{fig:topic-model1,fig:topic-model2} and \cref{fig:top-ten-terms}. In particular,
our method is capable of identifying dependencies in the topic space,
such as the cycle between topics that emerges from the crown motif.
In summary, the conceptual structure of $\ITT_{10}$ allows for a
global and deep investigation of the \SSHTM topic model.

\subsection{Zoom-In on topics}
\begin{figure}[!ht]
  \centering
  \scalebox{0.6}{\colorlet{mivertexcolor}{black!80}
\colorlet{jivertexcolor}{black!80}
\colorlet{vertexcolor}{black!80}
\colorlet{bordercolor}{black!80}
\colorlet{linecolor}{gray}
% parameter corresponds to the used valuation function and can be addressed by #1
\tikzset{vertexbase/.style 2 args={semithick, shape=circle, inner sep=2pt, outer sep=0pt, draw=bordercolor},%
  vertex/.style 2 args={vertexbase={#1}{}, fill=vertexcolor!45},%
  mivertex/.style 2 args={vertexbase={#1}{}, fill=mivertexcolor!45},%
  jivertex/.style 2 args={vertexbase={#1}{}, fill=jivertexcolor!45},%
  divertex/.style 2 args={vertexbase={#1}{}, top color=mivertexcolor!45, bottom color=jivertexcolor!45},%
  conn/.style={-, thick, color=linecolor}%
}
\tikzstyle{d} = [text width=3cm,align=center,label distance=1mm,execute at begin node=\setlength{\baselineskip}{0pt}]
\tikzstyle{f} = [text width=4cm,align=left,label distance=1mm,execute at begin node=\setlength{\baselineskip}{0pt}]
\tikzstyle{d2} = [text width=2cm,align=center,label distance=1mm,execute at begin node=\setlength{\baselineskip}{0pt}]
\tikzstyle{h} = [text width=4cm,align=center,label distance=0.5mm,execute at begin node=\setlength{\baselineskip}{0pt}]
\begin{tikzpicture}[scale=0.6,font=\normalsize,rotate=90]
  \begin{scope} %for scaling and the like
    \begin{scope} %draw vertices
      \foreach \nodename/\nodetype/\param/\xpos/\ypos in {%
        0/vertex//0./-5,
        1/vertex//6/-5,
        2/mivertex//-0/-1,
        3/vertex//5.5/-0.5,
        4/vertex//19/-0,
        5/vertex//-0/2,
        6/vertex//2/3,
        7/mivertex//9/3,
        8/mivertex//16/2.5,
        9/jivertex//5.5/3.5,
        10/vertex//-1.5/5,
        11/mivertex//16/5,
        12/vertex//20.5/5,
        13/divertex//-10/6,
        14/mivertex//-0/6,
        15/divertex//-7/6,
        16/mivertex//-4/6,
        17/mivertex//-12/6,
        18/mivertex//2/6,
        19/mivertex//9/6,
        20/mivertex//13/6,
        21/mivertex//22/6,
        22/mivertex//18/6,
        23/mivertex//26/6,
        24/vertex//5.5/9
      } \node[\nodetype={\param}{}] (\nodename) at (\xpos, \ypos) {};
    \end{scope}
    \begin{scope} %draw connections
      \path (12) edge[conn] (22);
      \path (6) edge[conn] (18);
      \path (21) edge[conn] (24);
      \path (4) edge[conn] (8);
      \path (9) edge[conn] (18);
      \path (3) edge[conn] (9);
      \path (22) edge[conn] (24);
      \path (5) edge[conn] (14);
      \path (7) edge[conn] (19);
      \path (1) edge[conn,in=-90] (8);
      \path (18) edge[conn] (24);
      \path (0) edge[conn,out=30,in=-90] (20);
      \path (13) edge[conn] (24);
      \path (4) edge[conn] (21);
      \path (0) edge[conn,in=-90,out=160] (13);
      \path (8) edge[conn] (20);
      \path (17) edge[conn] (24);
      \path (12) edge[conn] (21);
      \path (14) edge[conn] (24);
      \path (16) edge[conn] (24);
      \path (3) edge[conn] (7);
      \path (10) edge[conn] (16);
      \path (23) edge[conn] (24);
      \path (2) edge[conn] (5);
      \path (9) edge[conn] (19);
      \path (20) edge[conn] (24);
      \path (3) edge[conn] (5);
      \path (5) edge[conn] (15);
      \path (0) edge[conn,out=60,in=-90] (7);
      \path (11) edge[conn] (22);
      \path (1) edge[conn,in=-90,out=120] (6);
      \path (19) edge[conn] (24);
      \path (6) edge[conn] (14);
      \path (0) edge[conn,in=-90,out=120] (16);
      \path (3) edge[conn] (6);
      \path (10) edge[conn] (14);
      \path (15) edge[conn] (24);
      \end{scope}
    \begin{scope} %add labels
      \foreach \nodename/\labelpos/\labelopts/\labelcontent in {%
        2/above/d/{Kernel Methods},
        7/above/d/{Optimization},
        8/above right/f/{Feature Extraction, Support Vector Machines},
        11/above/d/{Social Media},
        13/above/d/{Graphs},
        14/above/d/{Matrix Methods},
        15/above/d/{Bayesian Inference},
        16/above/d/{Classification, Mining},
        17/above/d/{Semantic Web},
        18/above/d/{Learning, Knowledge Bases},
        19/above/d/{Reinforcement Learning},
        20/above/d/{Clustering},
        21/above/d/{Nonlinear Control},
        22/above/d/{Neurons, Dynamic Networks},
        23/above/d/{Image Recognition},
        24/above/h/{Neural Networks},
        0/below/d2/{\small algorithm},
        1/below/d2/{\small method},
        2/below/d2/{\small distribut},
        3/below/d2/{\small learn},
        4/below/d2/{\small propos},
        9/below/d2/{\small state},
        10/below/d2/{\small train},
        11/below/d2/{\small network},
        12/below/d2/{\small neural},
        13/below/d2/{\small structur, node},
        15/below/d2/{\small hidden},
        16/below/d2/{\small perform},
        17/below/d2/{\small link},
        22/below/d2/{\small recurr},
        23/below/d2/{\small deep, convolut},
        24/below/h/{\small feedforward, predict, architectur, backpropag, unit, local, layer, weight}
      } \coordinate[label={[\labelopts]\labelpos:{\labelcontent}}](c) at (\nodename);
    \end{scope}
  \end{scope}
\end{tikzpicture}}
  \caption{Zoom in on the \emph{Neural Network} topic of the
    term-topic relation for the \SSHTM topic model. The top term
    parameter is set to thirty.}
  \label{fig:topic-model-nn}
\end{figure}
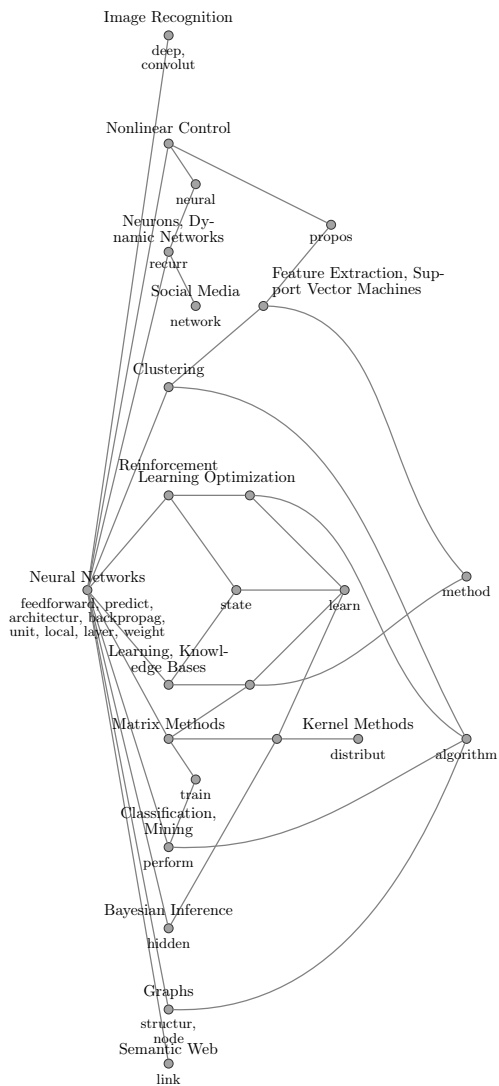

The limitation of $n=10$ can be softened by focussing on particular
topics of interest. We call this \emph{zoom-in on topics}. For
example, in our experiment, we are interested in concepts on
\emph{Neural Networks}. With the same reasoning as in the last
section, i.e., small number of concepts, we found $n=30$ to be
appropriate.  The direct approach to compute all concepts that contain
\emph{Neural Networks} is first compute all concepts of $\ITT_{30}$
and consecutively filter them for the topic in question.  In
\cref{fig:topic-model-nn} we depicted the result. Again, we omitted
the unsupported bottom element.  Out of 157 concepts do twenty-six
include the \emph{Neural Network} topic.  

We see in the diagram that some topics are drawn as lower neighbors of
other topics.  Restricted to the zoom-in on \emph{Neural Networks},
our method identifies several implications.  For example topic terms
of \textit{Kernel Methods} are also topic terms of \textit{Matrix
  Methods}. Another example is that topic terms of \emph{Optimization}
are also topic terms of \emph{Reinforcement Learning}. At this point
we want to note two important points about the interpretation of these
implications. First, the computed implications are valid within the
analyzed topic model. Hence, any logical conflicts with respect to
real world observations (or expert assessment) may indicate flaws of
the topic model. Second, for any implication the inverse is not
necessarily true.

We can identify several ordinal motifs in the zoomed-in structure.
For example, (1)
\emph{Classification}, \emph{Matrix Methods} and \emph{Optimization},
and (2) \emph{Clustering}, \emph{Learning Knowledge Base} and
\emph{Optimization}. Both are contranominal ordinal motifs.
Synonyms are important for training and applying topic models.  Within
\emph{Neural Networks} we can draw from the structure questions, such
as: What differentiates the terms \emph{algorithm} and \emph{method}?
In which topics are they used as synonyms? The same questions can be
formulated for \emph{train} and \emph{learn}.

\section{The Geometric Structure}\label{chapter:topic:sec:geometric}
Important aspects of data and their interpretation are captured
through geometric properties. This is in particular true for incidence
geometries. The study of ordinal motifs allows for further geometric
interpretation of the sub-structures within the topic space.  In the
last section, we analyzed the topic model with singular ordinal motifs
at a time. The goal now is to employ the comprehensive geometric
structure, i.e., the set of all ordinal motifs.  The \emph{geometric
  structure} of a (contextual) data set is a multi-relational
hypergraph structure.  In this hypergraph every hyperedge relation
encodes one type of ordinal motif $\mathcal{S}_1,\dots,\mathcal{S}_n$.

\begin{definition}[Geometric Structure]
  For a context $\context\coloneqq (G,M,I)$ and families of ordinal
  motifs $\mathcal{S}_1,\dots,\mathcal{S}_n$ is the \emph{geometric
    structure} of $\context$ with respect to
 $\mathcal{S}_1,\dots,\mathcal{S}_n$ a multi-hypergraph $(M,E_1,\dots,E_n)$
  where
  $$E_i\coloneqq \{N\subseteq M\mid \text{$N$ is of ordinal motif type $\mathcal{S}_{i}$ in $\context$}\}.$$
%  $$E_i\coloneqq \{H\subseteq G\mid \text{ex. a surjective local full \context[S]-measure $\sigma$ of }\context{[H,M]} \text{ with } \context[S]\in \mathcal{S}_{i}\}.$$
\end{definition}

An attribute set $N\subseteq M$ is of ordinal motif type
$\mathcal{S}$, if the concept lattice of $\context{[G,N]}$ (i.e.,
$\context$ restricted to the attributes $N$) resembles a concept
lattice of $\mathcal{S}$. For example, $\context{[G,N]}$ is of crown
type iff the concept lattice $\BV(\context{[G,N]})$ has a fence
pattern (cf. \cref{fig:substructures}).  We may note that we do not
impose any particular choice of $\mathcal{S}_i$.  The selection of
suitable ordinal motifs~\cite{ordinal-motif} is up to the analyst.
The question \emph{``Is $\context[G,N]$ of ordinal motif type?''} is a
formal decision problem called the \emph{ordinal motif
  problem}\footnote{Formally, we employ the surjective local full
  scale-measure ordinal motif problem. In the present work, we study
  ordinal motifs with respect to attribute sets. Therefor, all notions
  are translated to their dual counterparts with respect to the formal
  context.}. The corresponding computational complexities were
investigated in Hirth et al.~\cite{ordinal-motif}. All instances of
the ordinal motif problem that we investigate in this work are in
$\Pclass$ \cite{ordinal-motif-covering}.

\subsection{Geometric Structure Diagram}
To make the geometric structure human accessible, we decided for a
diagrammatic presentation. We define a set of drawing rules for each
ordinal motif type. After that, we apply our method to the \SSHTM
topic model. We call the resulting figure the \emph{geometric drawing}
of $\context$. This method is inspired by the geometric
representations introduced by R.~Wille drawing concept lattices
\cite{wille-geometric}.

In this geometric drawing, every attribute $m\in M$ (i.e., topic for
\SSHTM) is represented by a node. The hyperedges of an ordinal motif
type are drawn as connections between nodes. Each ordinal motif type
has its own drawing style to ensure that they can be distinguished.
The connection lines are annotated by the objects $g\in G$ (i.e.,
terms for \SSHTM) that induce the respective ordinal motif.
This allows for deriving explanations of ordinal motifs both in terms
of the attributes they connect and the objects they entail.

We present the drawing rules for each type of ordinal motifs in
greater detail:

\begin{description}
\item[Nominal] There are two cases of nominal ordinal motifs that we
  distinguish with respect to the objects they entail. In case there
  is an object $g\in G$ that supports all attributes $N$, i.e.,
  $g\in N'$, we draw an edge between all pairs of attributes
  $n_1,n_2\in N$ and annotate the connecting lines by $g$.  Otherwise,
  no edge is drawn. A prototypical example for both cases is shown in
  \cref{fig:substructures} (top right).
\item[Crown] Crown ordinal motifs do not require their own drawing
  rule. Instead, they can be read from (closed) cycles of nominal
  ordinal motifs. Yet, two conditions need to be satisfied: (1)
  Objects must not occur more than once along a cycle. (2) No other
  edges, apart from the cycle edges connect cycle attributes.  A
  prototypical example for the crown ordinal motif on ten elements is
  depicted in \cref{fig:substructures} (middle right).
\item[Contranominal] A contranominal ordinal motif encodes that any
  subset $D\subseteq N$ is supported by a unique set of objects
  $H\subseteq G$. Therefor, they reflect a densely connected part
  within the conceptual structure. Our drawing shall reflect this by
  applying the drawing rule: contranominal hyperedges of size $n$ are
  drawn by a filled $n$-polygon that connects the attributes $N$.  The
  edge between two attributes $n_1,n_2\in N$ then annotated by all
  objects that they have in common, i.e., by the objects
  $\{n_1,n_2\}'$. Prototypical examples for contranominal ordinal
  motifs of size three and four can be found in
  \cref{fig:substructures} (bot right) and \cref{fig:other-motifs}
  (bot right).
\item[Ordinal] Ordinal motifs that are of ordinal\footnote{We remind
    the reader that \emph{ordinal motif} is a defined class of objects
    and ordinal type addresses a particular sub-class.} type encode
  rankings among the attributes. In such a motif, the greatest element
  subsumes all the incidences of a smaller attribute.  We reflect this
  in the diagram by the drawing rule: an attribute (node) is drawn
  such that it overlaps the next lower ranked attribute (node).  For
  this kind of motif, objects are annotated next to nodes. At each
  attribute (node) we annotate all objects such that there is no lower
  ranked attribute that is in incidence with this object.  This
  procedure is in accordance to the short-hand notation of FCA line
  diagrams. A prototypical example is depicted in
  \cref{fig:other-motifs} (top right).
\item[Interordinal] The interordinal ordinal motif encodes two ordinal
  motifs of ordinal type, whose rankings on the attributes are
  complementary\footnote{ A natural example for this are the
    \emph{``$x$ is hotter than $y$''} and \emph{``$y$ is colder than
      $x$''} relations.} to each other. We have depicted an example on
  four elements in \cref{fig:other-motifs} (middle right). To display
  an interordinal ordinal motif one should draw an hyperedge that
  encloses the motif's attributes. The objects are annotated next to
  the attribute nodes based on the two rankings and the ordinal
  drawing rule. Objects from the same ranking have to be drawn on the
  same side of the hyperedge.
\end{description}

\subsection{The Geometric Structure of \SSHTM}
\begin{figure}[t]
  \centering
  \includegraphics{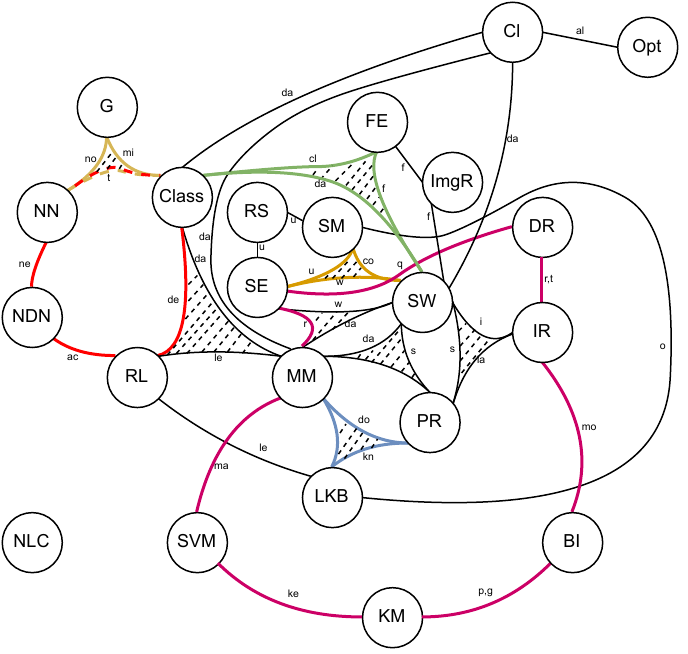}
  \caption{The geometric drawing of the \SSHTM topic model for top-ten
    terms (cf. \cref{fig:top-n-term-concepts}). Topic and term names
    are abbreviated for better readability. Their full length names can be 
    inferred from \cref{fig:top-ten-terms}.}
  % edges with the same label are the same subconcept
  \label{fig:geometric-structure}
\end{figure}

The resulting geometric drawing for \SSHTM is depict in
\cref{fig:geometric-structure}.  The drawing includes nominal,
contranominal and crown ordinal motifs. The are no non-trivial ordinal
or interordinal types in the topic model structure. This fact is,
however, not surprising, since the topics within a topic model are
optimized to be independent of each other. For readability reasons, we
abbreviated the terms and topics. Their un-abbreviated versions are
listed in \cref{fig:top-ten-terms}.

In order to increase the readability and comparability to the last
section (cf. \cref{fig:top-n-term-concepts}), we have highlighted
ordinal motifs in the same color.  The representation of the relations
in the geometric drawing is fundamentally different to line diagrams.
By design, in the geometric drawing it is easier to identify ordinal
motifs. For example, we can easily read from the diagram that there
are eight contranominal ordinal motifs. Out of the twenty-two topics,
\emph{Semantic Web} (SW) occurs in five contranominal ordinal motifs
and is therefore structurally very important for within the topic
model.  This is followed by the \emph{Matrix Methods} (MM) topic which
occurs in four contranominal ordinal motifs. In contrast to line
diagrams, this information is easy to infer from the geometric
drawing. The \emph{Non-Linear Control} (NLC) topic is very isolated
and does not exhibit any (non-trivial) connection to other topics.

Crown ordinal motifs can easily be read from the structure as (closed)
cycles. For example, we find \emph{NN -- Class -- RL -- NDN --NN}
(orange) and \emph{SW -- IR -- DR -- SE -- SW}. Both crowns identified
in the line diagram \cref{fig:top-n-term-concepts} are highlighted in
the geometric drawing in the same color.

We invite the reader to compare the geometric drawing to classical
approaches such as topic-topic heatmaps and t-SNE embeddings, as
depicted in \cref{fig:topic-model1,fig:topic-model2}. Based on this
comparison, we argue that geometric drawings of topic models allow for
a non-flat analysis of the inter-topic relation and their respective
terms.
\FloatBarrier
\section{Limitations \& Conclusion}
With our work, we proposed a comprehensive approach for analyzing and
visualizing high dimensional topic models. In principle, this method
is applicable to arbitrary matrix shaped data sets.  We have shown
that our method is capable of capturing insights about researchers and
venues from the realm of machine learning research. Moreover, we
demonstrated how conceptual structures can be used to track the change
in their topics. For our analysis, we employed ordinal patterns which
occurred frequently in the data. These sub-structures allow for a rich
interpretation of the topic model. In particular, the inter-topic and
term-topic relation.

This interpretability, of course, depends on the overall understanding
of the terms of the topic model. Hence, although our method is
applicable to arbitrary matrix shaped data sets, meaningful
interpretations are limited by the available background knowledge.
Another limitation of our method is the number of concepts one can
visualize in a readable fashion.  This number is dependent on the
number of topics, documents and selected top terms per topic.
To compensate for this limitation, we proposed the use of (graph) core
structures.

As the present work has established a robust link between topic models
and their conceptual analysis, we envision several directions for
future work. First, the absence of (non-trivial) ordinal and
interordinal motifs within the analyzed topic model is not
surprising. This due to the fact that topic models optimize to compute
independent (non-nesting) topics. Yet, this is not true in the case of
hierarchical topic modeling. The logical next step is to apply our
methods to these models, e.g., HLDA or PAM
\cite{FCA-TM2,zhao2018inter,hlda,PAM-HTM}.

Second, within the research field of human-computer interaction, we
propose to conduct a user study in order to gather statistical
evidence. Moreover, this may reveal new insights into the developed
geometric drawings and potentially their visual optimization.
Third, in order to conduct a study, as proposed above, a difficult
algorithmic task has to be solved. Although, the geometric drawings
are well-defined, their algorithmic computation is an open problem.

%\printbibliography

\bibliographystyle{elsarticle-num} 
\bibliography{paper.bib}

\appendix
\section{Appendix}
% \begin{figure}
%   \centering
%   \input{pics/fox.tikz}
%   \input{pics/thrun.tikz}
%   \input{pics/ecml.tikz}
%   \caption{Author View Fox (top) Thrun (Middle) ECML (bot)}
%   \label{fig:fox-schoelkopf-view}
% \end{figure}

\begin{figure}[H]
  \centering
  \includegraphics[height=\linewidth,rotate=90]{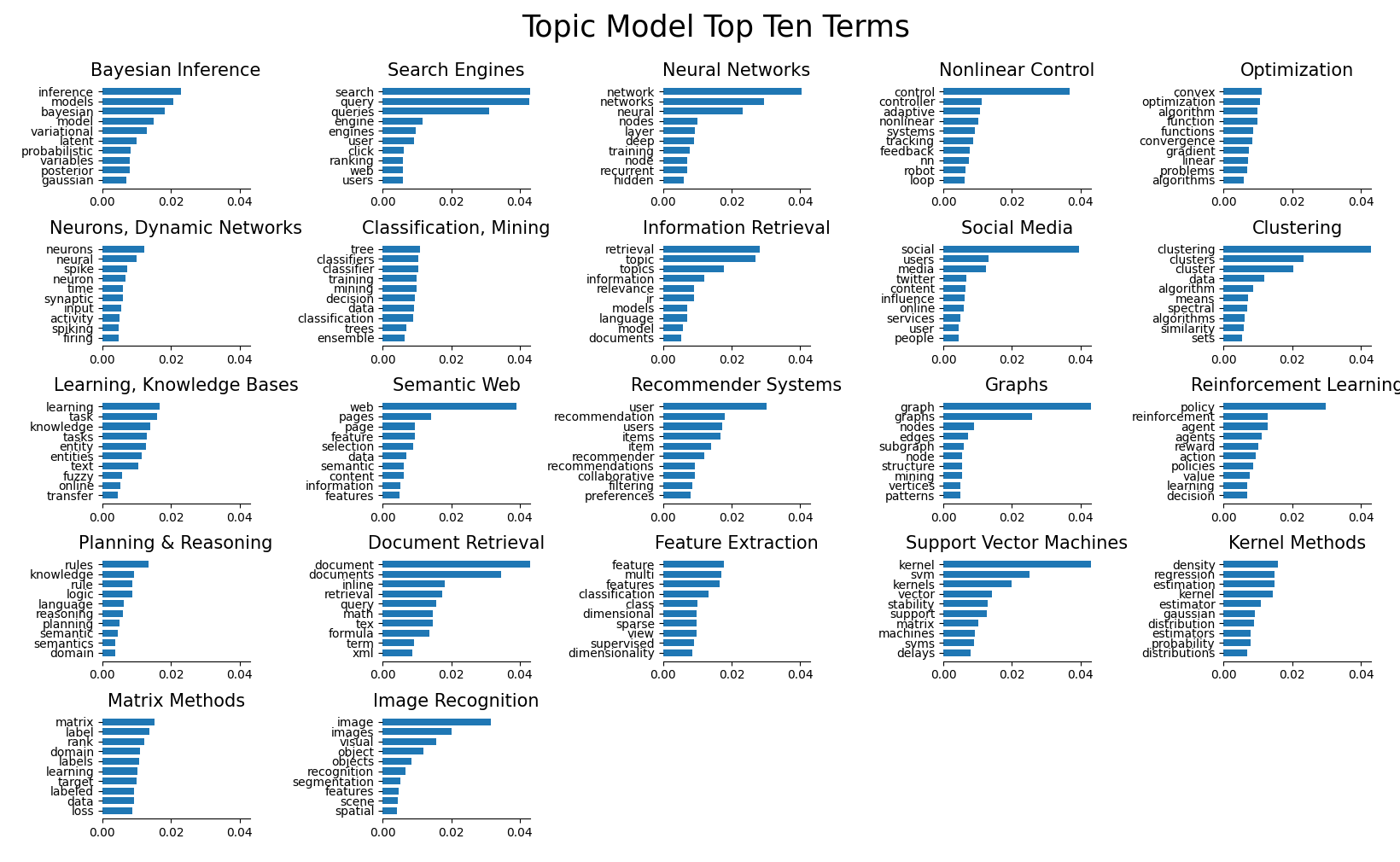}
  \caption{The top ten terms for each topic of the \SSHTM topic model \cite{TopicSpaceTrajectories}.}
  \label{fig:top-ten-terms}
\end{figure}

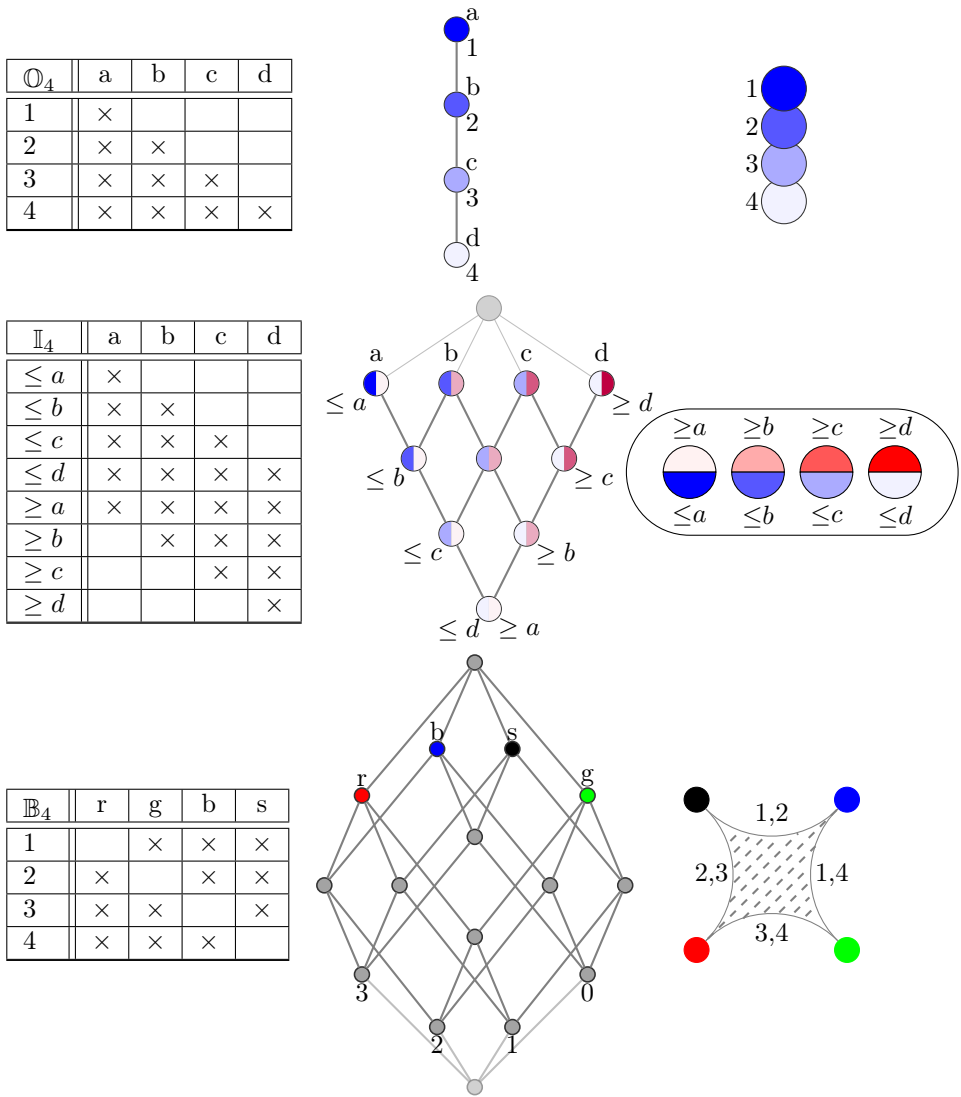
\begin{figure}[H]
  \centering
\begin{minipage}{0.33\linewidth}
    \begin{cxt} 
      \cxtName{$\mathbb{O}_{4}$}
      \att{a}
      \att{b}
      \att{c}
      \att{d}
      \obj{x...}{1}
      \obj{xx..}{2}
      \obj{xxx.}{3}
      \obj{xxxx}{4}
    \end{cxt}  
\end{minipage}
\begin{minipage}{0.33\linewidth}\centering
  \begin{tikzpicture}[concept/.style={fill=black!36,shape=circle,draw=black!80},
                      hidden/.style={opacity=0.5},
                      edge/.style={color=gray},]

    \node[concept,fill=blue!5] (a) at (0,0){};
    \node[concept,fill=blue!33] (b) at (0,1){};
    \node[concept,fill=blue!66] (c) at (0,2){};
    \node[concept,fill=blue!100] (d) at (0,3){};
    \path[edge,thick] (a) edge (b);
    \path[edge,thick] (b) edge (c);
    \path[edge,thick] (c) edge (d);
    \coordinate[label={above right:{a}}]() at (d);
    \coordinate[label={above right:{b}}]() at (c);
    \coordinate[label={above right:{c}}]() at (b);
    \coordinate[label={above right:{d}}]() at (a);

    \coordinate[label={below right:{1}}]() at (d);
    \coordinate[label={below right:{2}}]() at (c);
    \coordinate[label={below right:{3}}]() at (b);
    \coordinate[label={below right:{4}}]() at (a);

  \end{tikzpicture}
\end{minipage}
\begin{minipage}{0.32\linewidth}\centering
  \begin{tikzpicture}[concept/.style={fill=black!36,shape=circle,draw=black!80},
                      hidden/.style={opacity=0.5},
                      edge/.style={color=gray},]

    \node[concept,fill=blue!5,inner sep=6pt] (a) at (0,0){};
    \node[concept,fill=blue!33,inner sep=6pt] (b) at (0,0.5){};
    \node[concept,fill=blue!66,inner sep=6pt] (c) at (0,1){};
    \node[concept,fill=blue!100,inner sep=6pt] (d) at (0,1.5){};

    \coordinate[label={[label distance=6pt]left:{4}}]() at (a);
    \coordinate[label={[label distance=6pt]left:{3}}]() at (b);
    \coordinate[label={[label distance=6pt]left:{2}}]() at (c);
    \coordinate[label={[label distance=6pt]left:{1}}]() at (d);
  \end{tikzpicture}
\end{minipage}

%%% Local Variables:
%%% mode: latex
%%% TeX-master: "../paper"
%%% End:

\, \\

\begin{minipage}{0.33\linewidth}
    \begin{cxt} 
      \cxtName{$\mathbb{I}_{4}$}
      \att{a}
      \att{b}
      \att{c}
      \att{d}
      \obj{x...}{$\leq a$}
      \obj{xx..}{$\leq b$}
      \obj{xxx.}{$\leq c$}
      \obj{xxxx}{$\leq d$}
      \obj{xxxx}{$\geq a$}
      \obj{.xxx}{$\geq b$}
      \obj{..xx}{$\geq c$}
      \obj{...x}{$\geq d$}
    \end{cxt}  
\end{minipage}
\begin{minipage}{0.33\linewidth}\centering
  \begin{tikzpicture}[concept/.style={fill=black!36,shape=circle,draw=black!80},
                      hidden/.style={opacity=0.5},
                      edge/.style={color=gray},rotate=180]

    \node[concept,circle split part fill={blue!5, purple!5}] (d) at (0,5){};

    \node[concept,circle split part fill={blue!5, purple!33}] (c) at (-0.5,4){};
    \node[concept,circle split part fill={blue!33, purple!5}] (ad) at (0.5,4){};

    \node[concept,circle split part fill={blue!5, purple!66}] (b) at (-1,3){};
    \node[concept,circle split part fill={blue!33, purple!33}] (m1) at (0,3){};
    \node[concept,circle split part fill={blue!66, purple!5}] (bd) at (1,3){};

    \node[concept,circle split part fill={blue!5, purple!100}] (a) at (-1.5,2){};
    \node[concept,circle split part fill={blue!33, purple!66}] (m2) at (-0.5,2){};
    \node[concept,circle split part fill={blue!66, purple!33}] (m3) at (0.5,2){};
    \node[concept,circle split part fill={blue!100, purple!5}] (cd) at (1.5,2){};

    \node[concept,hidden] (bot) at (0,1){};

    \path[edge,hidden] (a) edge (bot);
    \path[edge,hidden] (m2) edge (bot);
    \path[edge,hidden] (m3) edge (bot);
    \path[edge,hidden] (cd) edge (bot);

    \path[edge,thick] (b) edge (a);
    \path[edge,thick] (b) edge (m2);
    \path[edge,thick] (m1) edge (m2);
    \path[edge,thick] (m1) edge (m3);
    \path[edge,thick] (bd) edge (m3); 
    \path[edge,thick] (bd) edge (cd); 

    \path[edge,thick] (c) edge (b);
    \path[edge,thick] (c) edge (m1);
    \path[edge,thick] (ad) edge (m1);
    \path[edge,thick] (ad) edge (bd);

    \path[edge,thick] (d) edge (c);
    \path[edge,thick] (d) edge (ad);

    \coordinate[label={below right:{$\geq d$}}]() at (a);
    \coordinate[label={below right:{$\geq c$}}]() at (b);
    \coordinate[label={below right:{$\geq b$}}]() at (c);
    \coordinate[label={below right:{$\geq a$}}]() at (d);

    \coordinate[label={below left:{$\leq d$}}]() at (d);
    \coordinate[label={below left:{$\leq c$}}]() at (ad);
    \coordinate[label={below left:{$\leq b$}}]() at (bd);
    \coordinate[label={below left:{$\leq a$}}]() at (cd);

    \coordinate[label={[label distance=5pt]above:{d}}]() at (a);
    \coordinate[label={[label distance=5pt]above:{c}}]() at (m2);
    \coordinate[label={[label distance=5pt]above:{b}}]() at (m3);
    \coordinate[label={[label distance=5pt]above:{a}}]() at (cd);

  \end{tikzpicture}
\end{minipage}
\begin{minipage}{0.32\linewidth}\centering

  \begin{tikzpicture}[concept/.style={fill=black!36,shape=circle,draw=black!80},
                      hidden/.style={opacity=0.5},
                      edge/.style={color=gray},scale=0.7]
    % \node[concept,circle split part fill={blue!5, purple!5},inner sep=6pt] (d) at (0,0){};
    % \node[concept,circle split part fill={blue!33, purple!33},inner sep=6pt] (c) at (0,0.5){};
    % \node[concept,circle split part fill={blue!66, purple!66},inner sep=6pt] (b) at (0,1){};
    % \node[concept,circle split part fill={blue!100, purple!100},inner sep=6pt] (a) at (0,1.5){};

    % \coordinate[label={[label distance=6pt]right:{4}}]() at (a);
    % \coordinate[label={[label distance=6pt]right:{3}}]() at (b);
    % \coordinate[label={[label distance=6pt]right:{2}}]() at (c);
    % \coordinate[label={[label distance=6pt]right:{1}}]() at (d);

    % \coordinate[label={[label distance=6pt]left:{1}}]() at (a);
    % \coordinate[label={[label distance=6pt]left:{2}}]() at (b);
    % \coordinate[label={[label distance=6pt]left:{3}}]() at (c);
    % \coordinate[label={[label distance=6pt]left:{4}}]() at (d);

    % red order
     \draw [fill=red!5] (0,0) to[out=90,in=180] (0.5,0.5) to[out=0,in=90] (1,0) -- cycle;
     \draw [fill=red!33] (1.3,0) to[out=90,in=180] (1.8,0.5) to[out=0,in=90] (2.3,0) -- cycle;
     \draw [fill=red!66] (2.6,0) to[out=90,in=180] (3.1,0.5) to[out=0,in=90] (3.6,0) -- cycle;
     \draw [fill=red!100] (3.9,0) to[out=90,in=180] (4.4,0.5) to[out=0,in=90] (4.9,0) -- cycle;

    % blue order
    \draw [fill=blue!5] (3.9,0) to[out=-90,in=180] (4.4,-0.5) to[out=0,in=-90] (4.9,0) -- cycle;
    \draw [fill=blue!33] (2.6,0) to[out=-90,in=180] (3.1,-0.5) to[out=0,in=-90] (3.6,0) -- cycle;
    \draw [fill=blue!66] (1.3,0) to[out=-90,in=180] (1.8,-0.5) to[out=0,in=-90] (2.3,0) -- cycle;
    \draw [fill=blue!100] (0,0) to[out=-90,in=180] (0.5,-0.5) to[out=0,in=-90] (1,0) -- cycle;

    \draw (-0.7,0) to[out=90,in=180] (0.5,1.2) to (4.4,1.2) to[out=0,in=90] (5.6,0) to[out=-90,in=0] (4.6,-1.2) to (0.5,-1.2) to[out=180,in=-90] (-0.7,0);

    \coordinate (a) at (0.5,0);
    \coordinate (b) at (1.8,0);
    \coordinate (c) at (3.1,0);
    \coordinate (d) at (4.4,0);

    \coordinate[label={[label distance=9pt]below:{${\leq} a$}}]() at (a);
    \coordinate[label={[label distance=9pt]below:{${\leq} b$}}]() at (b);
    \coordinate[label={[label distance=9pt]below:{${\leq} c$}}]() at (c);
    \coordinate[label={[label distance=9pt]below:{${\leq} d$}}]() at (d);

    \coordinate[label={[label distance=9pt]above:{${\geq} a$}}]() at (a);
    \coordinate[label={[label distance=9pt]above:{${\geq} b$}}]() at (b);
    \coordinate[label={[label distance=9pt]above:{${\geq} c$}}]() at (c);
    \coordinate[label={[label distance=9pt]above:{${\geq} d$}}]() at (d);
  \end{tikzpicture}
\end{minipage}

%%% Local Variables:
%%% mode: latex
%%% TeX-master: "../paper"
%%% End:

\, \\

  \begin{minipage}{0.33\linewidth}
    \begin{cxt} 
      \cxtName{$\mathbb{B}_{4}$}
      \att{r}
      \att{g}
      \att{b}
      \att{s}
      \obj{.xxx}{1}
      \obj{x.xx}{2}
      \obj{xx.x}{3}
      \obj{xxx.}{4}
    \end{cxt}  
  \end{minipage}
  \begin{minipage}{0.33\linewidth}\centering
    \colorlet{mivertexcolor}{black!80}
\colorlet{jivertexcolor}{black!80}
\colorlet{vertexcolor}{black!80}
\colorlet{bordercolor}{black!80}
\colorlet{linecolor}{gray}
% parameter corresponds to the used valuation function and can be addressed by #1
\tikzset{vertexbase/.style 2 args={semithick, shape=circle, inner sep=2pt, outer sep=0pt, draw=bordercolor},%
  vertex/.style 2 args={vertexbase={#1}{}, fill=vertexcolor!45},%
  rvertex/.style 2 args={vertexbase={#1}{}, fill=red},%
  gvertex/.style 2 args={vertexbase={#1}{}, fill=green},%
  bvertex/.style 2 args={vertexbase={#1}{}, fill=blue},%
  svertex/.style 2 args={vertexbase={#1}{}, fill=black},%
  mivertex/.style 2 args={vertexbase={#1}{},opacity=0.5, fill=mivertexcolor!45},%
  jivertex/.style 2 args={vertexbase={#1}{}, fill=jivertexcolor!45},%
  divertex/.style 2 args={vertexbase={#1}{}, top color=mivertexcolor!45, bottom color=jivertexcolor!45},%
  conn/.style={-, thick, color=linecolor}%
}
\begin{tikzpicture}
  \begin{scope} %for scaling and the like
    \begin{scope} %draw vertices
      \foreach \nodename/\nodetype/\param/\xpos/\ypos in {%
        15/mivertex//0/-0,
        0/vertex//-0.5/0.8,
        1/vertex//0.5/0.8,
        2/vertex//-1.5/1.5,
        3/vertex//1.5/1.5,
        4/vertex//0/2,
        5/vertex//-2/2.685,
        6/vertex//-1/2.685,
        7/vertex//1/2.685,
        8/vertex//2/2.685,
        9/vertex//0/3.33,
        10/rvertex//-1.5/3.88,
        11/gvertex//1.5/3.88,
        12/bvertex//-0.5/4.5,
        13/svertex//0.5/4.5,
        14/vertex//0/5.65
      } \node[\nodetype={\param}{}] (\nodename) at (\xpos, \ypos) {};
    \end{scope}
    \begin{scope} %draw connections
      \path (6) edge[conn,thick] (13);
      \path (6) edge[conn,thick] (10);
      \path (9) edge[conn,thick] (13);
      \path (9) edge[conn,thick] (12);
      \path (4) edge[conn,thick] (11);
      \path (4) edge[conn,thick] (10);
      \path (1) edge[conn,thick] (6);
      \path (1) edge[conn,thick] (4);
      \path (1) edge[conn,thick] (8);
      \path (11) edge[conn,thick] (14);
      \path (13) edge[conn,thick] (14);
      \path (5) edge[conn,thick] (12);
      \path (5) edge[conn,thick] (10);
      \path (8) edge[conn,thick] (11);
      \path (8) edge[conn,thick] (13);
      \path (12) edge[conn,thick] (14);
      \path (10) edge[conn,thick] (14);
      \path (0) edge[conn,thick] (4);
      \path (0) edge[conn,thick] (5);
      \path (0) edge[conn,thick] (7);
      \path (2) edge[conn,thick] (6);
      \path (2) edge[conn,thick] (9);
      \path (2) edge[conn,thick] (5);
      \path (3) edge[conn,thick] (9);
      \path (3) edge[conn,thick] (8);
      \path (3) edge[conn,thick] (7);
      \path (7) edge[conn,thick] (11);
      \path (7) edge[conn,thick] (12);
      \path (0) edge[conn,opacity=0.5] (15);
      \path (1) edge[conn,opacity=0.5] (15);
      \path (2) edge[conn,opacity=0.5] (15);
      \path (3) edge[conn,opacity=0.5] (15);
    \end{scope}
    \begin{scope} %add labels
      \foreach \nodename/\labelpos/\labelopts/\labelcontent in {%
        10/above//{r},
        11/above//{g},
        12/above//{b},
        13/above//{s},
        0/below//{2},
        1/below//{1},
        2/below//{3},
        3/below//{0}
      } \coordinate[label={[\labelopts]\labelpos:{\labelcontent}}](c) at (\nodename);
    \end{scope}
  \end{scope}
\end{tikzpicture}
  \end{minipage}
  \begin{minipage}{0.32\linewidth}\centering
      \begin{tikzpicture}[concept/.style={fill=black!36,shape=circle,draw=black!80},
                      hidden/.style={opacity=0.5},
                      edge/.style={color=gray},]

    \draw[edge,dashed,thick] (0,0) -- (2,2);
    \draw[edge,dashed,thick] (0.62,0.42) -- (1.58,1.38);
    \draw[edge,dashed,thick] (0.9,0.5) -- (1.45,1.05);
    \draw[edge,dashed,thick] (1.1,0.5) -- (1.5,0.9);
    \draw[edge,dashed,thick] (1.3,0.5) -- (1.5,0.7);
    \draw[edge,dashed,thick] (1.45,0.45) -- (1.55,0.55);
    \draw[edge,dashed,thick] (0.44,0.64) -- (1.35,1.55);
    \draw[edge,dashed,thick] (0.5,0.9) -- (1.05,1.45);
    \draw[edge,dashed,thick] (0.5,1.1) -- (0.9,1.5);
    \draw[edge,dashed,thick] (0.5,1.3) -- (0.7,1.5);
    \draw[edge,dashed,thick] (0.45,1.45) -- (0.55,1.55);

    \node[concept,color=red] (a) at (0,0){};
    \node[concept,color=green] (b) at (2,0){};
    \node[concept,color=blue] (c) at (2,2){};
    \node[concept,color=black] (d) at (0,2){};

    \node at (0.2,1){2,3};
    \node at (1,0.2){3,4};
    \node at (1,1.8){1,2};
    \node at (1.8,1){1,4};

    \path[edge,in=135,out=45] (a) edge (b);
    \path[edge,out=135,in=225] (b) edge (c);
    \path[edge,out=225,in=-45] (c) edge (d);
    \path[edge,out=-45,in=45] (d) edge (a);
    % \coordinate[label={below left:{a}}]() at (a);
    % \coordinate[label={below right:{b}}]() at (b);
    % \coordinate[label={above right:{c}}]() at (c);
    % \coordinate[label={above left:{d}}]() at (d);

  \end{tikzpicture}
  \end{minipage}

%%% Local Variables:
%%% mode: latex
%%% TeX-master: "../paper"
%%% End:

\caption{The formal context, concept lattice and geometric drawing style of the ordinal, interordinal and contranominal ordinal motif.}
  \label{fig:other-motifs}
\end{figure}

\end{document}